%% file: phyai.tex
\documentclass[10pt, a4paper, nonumbering]{phyai}

\usepackage{natbib}
\setcitestyle{authoryear,round,citesep={;},aysep={,},yysep={;}}
\usepackage{dblfloatfix}
\usepackage[normalem]{ulem}
\usepackage{caption}
\usepackage{dramatist}
\usepackage{xspace}
\usepackage{pifont}
\usepackage{multirow}
\usepackage{tcolorbox}
\usepackage{xltabular}
\usepackage{longtable}
\usepackage{hyperref}
\hypersetup{
    colorlinks=false,
    pdfborder={0 0 1},
    linkbordercolor={1 0 0},
    citebordercolor={0 1 0},
    urlbordercolor={1 0 0}
}
\usepackage{wrapfig}
\usepackage{graphicx}
\usepackage[inkscapelatex=false,inkscapeopt=--export-text-to-path]{svg}
\usepackage{placeins}

\usepackage{amsfonts}
\usepackage{amsmath}
\usepackage{amssymb}
\usepackage{lineno}
\usepackage{multirow}
\usepackage{adjustbox}

\usepackage[bottom]{footmisc}
\usepackage{algorithm}
\usepackage{float}
\usepackage{algpseudocode}
\usepackage{CJKutf8}
\usepackage{subcaption}
\usepackage{setspace}
\usepackage{makecell}
\usepackage{graphicx}
\usepackage{multicol}
\usepackage{xspace}
\usepackage{cleveref}

\algnewcommand{\Initialize}{\textbf{Initialize: }}

\definecolor{phyaiblue}{HTML}{0756C9}
\definecolor{phyaideep}{HTML}{172B3F}
\definecolor{phyaipale}{HTML}{DCE7F2}
\definecolor{phyaitodo}{HTML}{16833A}

\usepackage{fontspec}
\newcolumntype{L}[1]{>{\raggedright\arraybackslash}p{#1}}
\newcolumntype{Y}{>{\raggedright\arraybackslash}X}
\setlist[itemize,enumerate]{topsep=3pt,itemsep=2pt,parsep=0pt,partopsep=0pt}
\titleformat{\section}
    {\sffamily\bfseries\Large\color{phyaiblue}}
    {\thesection}
    {0.65em}
    {#1}
    [\vspace{2pt}\color{phyaiblue}\titlerule]
\titleformat{name=\section,numberless}
    {\sffamily\bfseries\Large\color{phyaiblue}}
    {}
    {0pt}
    {#1}
    [\vspace{2pt}\color{phyaiblue}\titlerule]
\titleformat{\subsection}
    {\sffamily\bfseries\large\color{phyaiblue}}
    {\thesubsection}
    {0.65em}
    {#1}
\titleformat{\subsubsection}
    {\sffamily\bfseries\normalsize\color{phyaiblue}}
    {\thesubsubsection}
    {0.65em}
    {#1}
\titleformat{\paragraph}[runin]
    {\sffamily\bfseries\normalsize\color{phyaiblue}}
    {\theparagraph}
    {0.65em}
    {#1}
\titlespacing*{\section}{0pt}{2.1ex plus .5ex minus .4ex}{0.8ex}
\titlespacing*{\subsection}{0pt}{1.7ex plus .4ex minus .3ex}{0.45ex}
\titlespacing*{\subsubsection}{0pt}{1.4ex plus .3ex minus .2ex}{0.35ex}
\titlespacing{\paragraph}{0pt}{1.2ex plus .3ex minus .2ex}{0.55em}

\begin{abstract}
    \input{00-abs}
\end{abstract}

\renewcommand{\abscontent}{%
    \noindent
    {\sffamily\bfseries\fontsize{9.5pt}{11pt}\selectfont Abstract}\vspace{0.3ex}\par
    \noindent\parbox{\linewidth}{\normalfont\fontsize{8.4pt}{9.8pt}\selectfont \theabstract}%
}

\begin{document}
{
\bgroup
\setlength{\parindent}{0pt}
\vspace*{-26pt}
\begin{adjustwidth}{0pt}{0pt}
{\raggedright
\noindent
\begin{minipage}[c]{0.34\linewidth}
    \includegraphics[width=88pt]{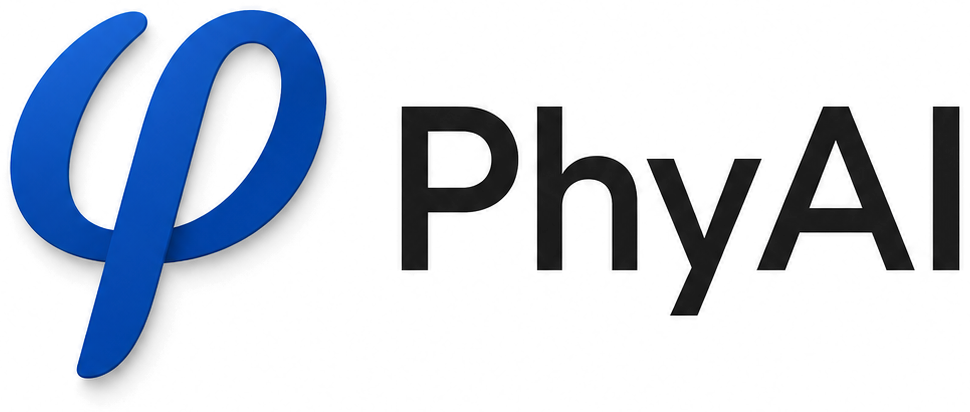}
\end{minipage}%
\hfill
\begin{minipage}[c]{0.62\linewidth}
    \raggedleft
    \includegraphics[height=28pt]{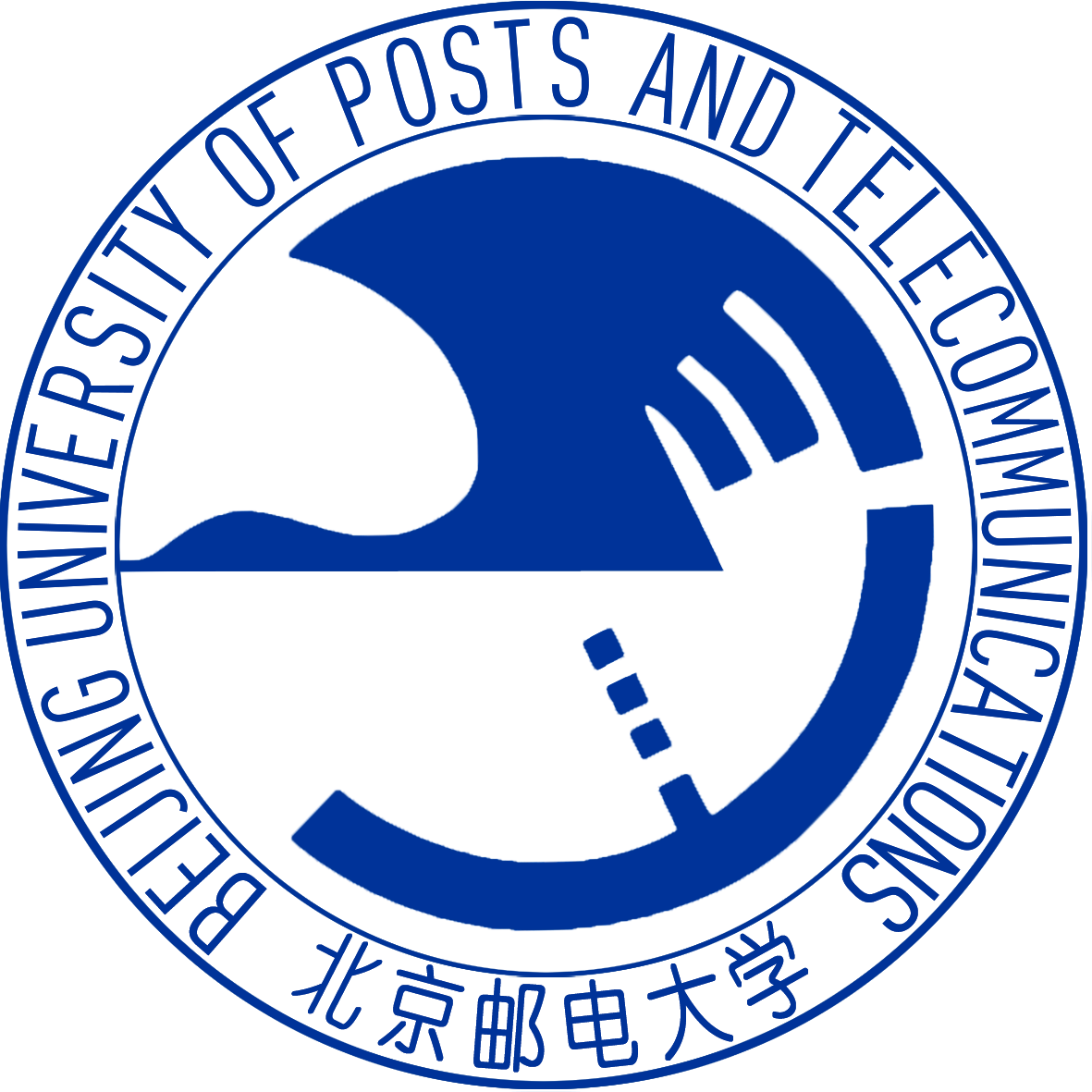}\hspace{7pt}%
    \includegraphics[height=28pt]{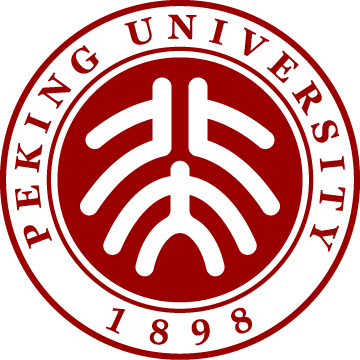}\hspace{7pt}%
    \includegraphics[height=28pt]{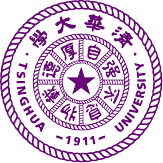}\hspace{7pt}%
    \includegraphics[height=28pt]{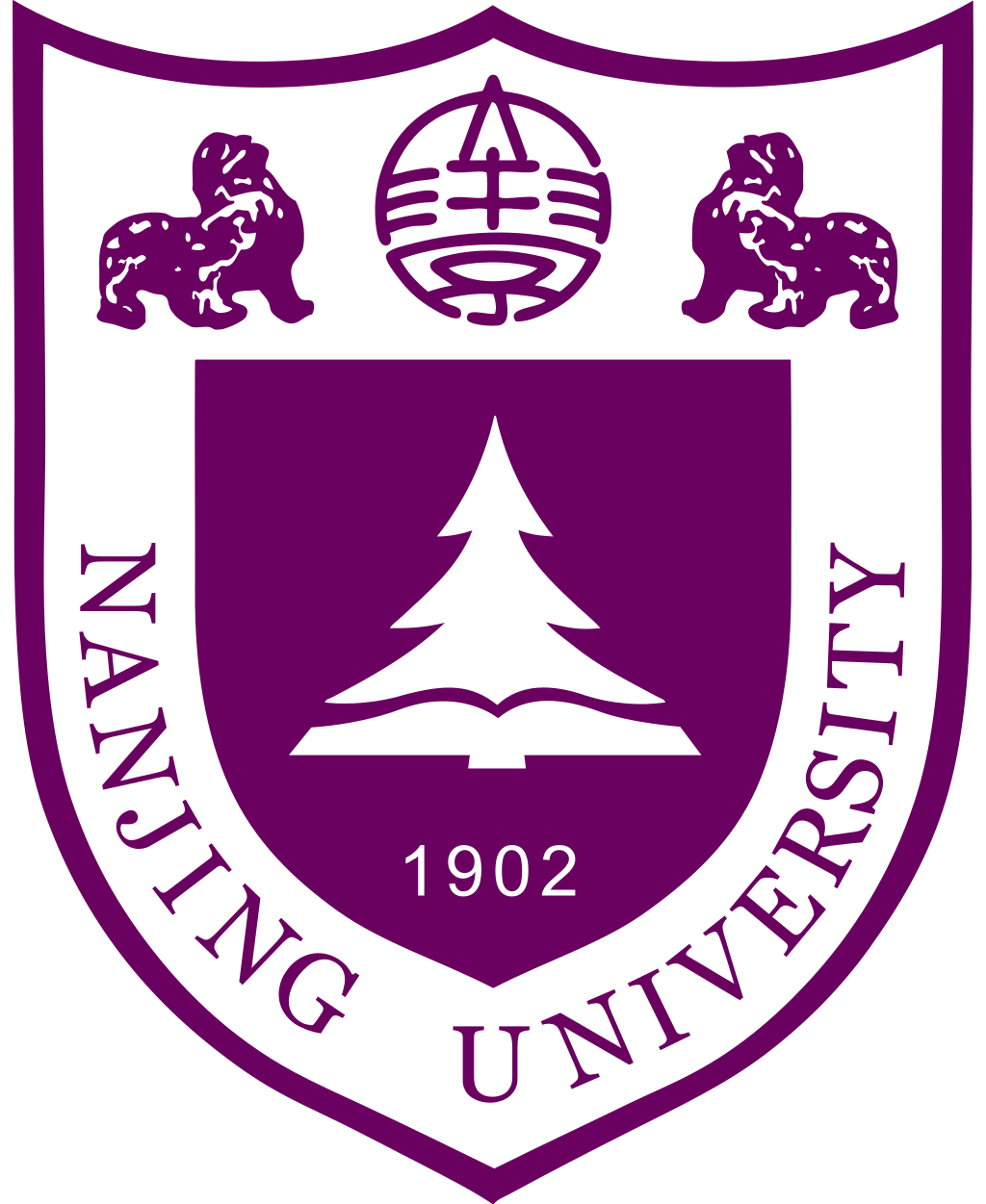}\hspace{9pt}%
    \raisebox{1pt}{\includegraphics[height=26pt,trim=250 180 240 110,clip]{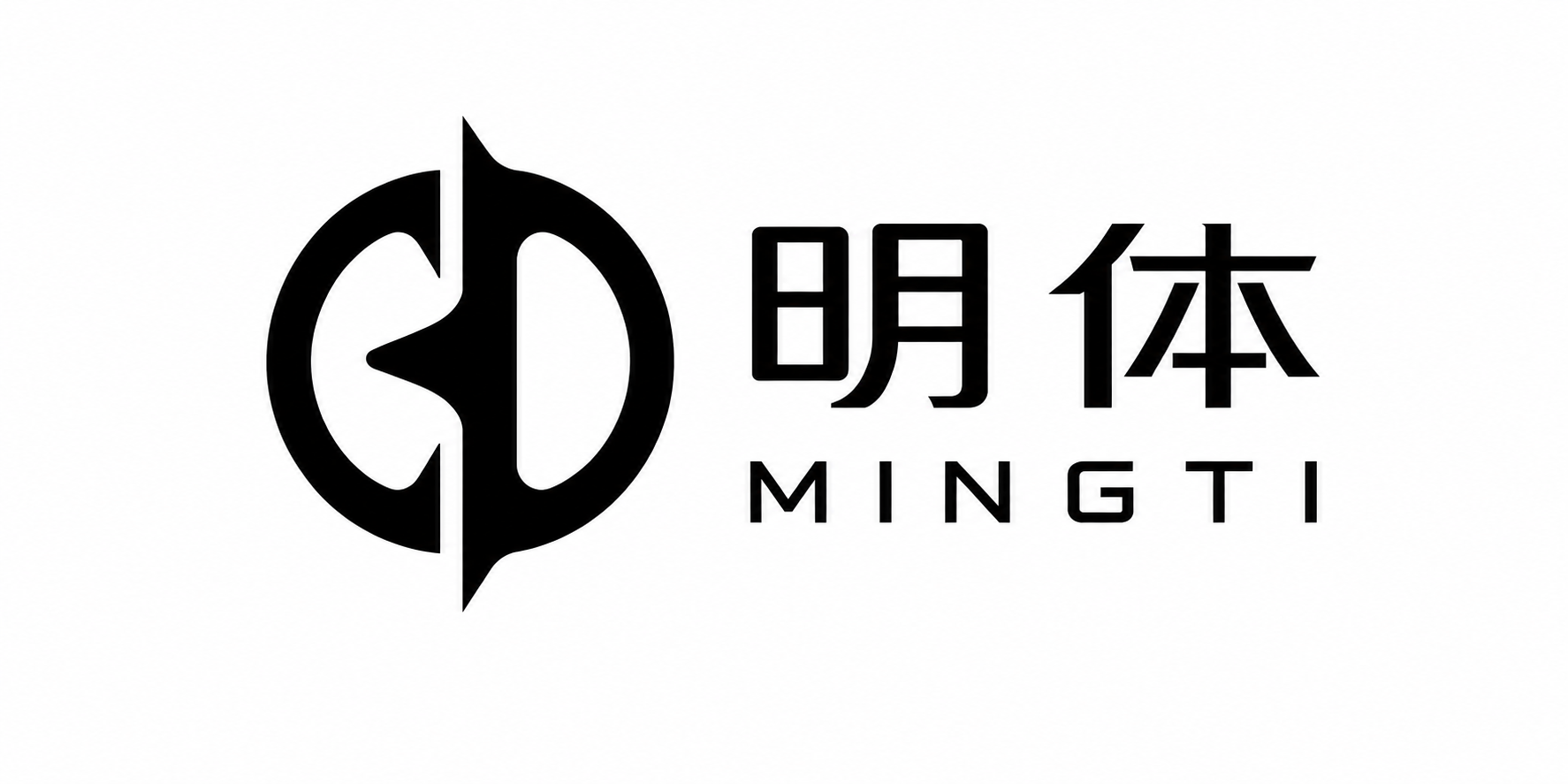}}\hspace{8pt}%
    \raisebox{5.5pt}{\includegraphics[height=17pt]{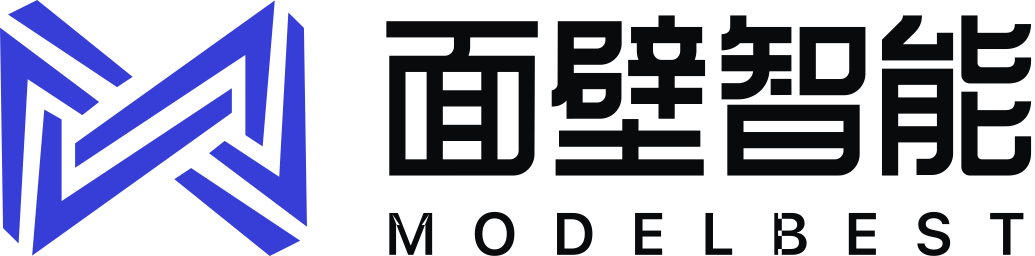}}
\end{minipage}
\par\vspace{1pt}
{\color{phyaiblue}\rule{\linewidth}{0.9pt}\par}
\vspace{9pt}
    {\sffamily\bfseries\fontsize{19}{22}\selectfont\color{phyaiblue} PhyAI: Real-Time Physical AI at the Edge,\\
Scalable Rollouts in the Cloud\par}
{
\vskip7pt
    {\sffamily\fontsize{8.5}{11.4}\selectfont
        \mbox{Chenghua Wang\textsuperscript{\textdagger,1}} \quad \mbox{Daliang Xu\textsuperscript{*,1}} \quad \mbox{Dongqi Cai\textsuperscript{2}} \quad \mbox{Duojin Sun\textsuperscript{1}} \quad \mbox{Hao Zhang\textsuperscript{1}} \quad \mbox{Haoze Qian\textsuperscript{1}} \quad \mbox{Huaiyuan Zhang\textsuperscript{1}}\\
        \mbox{Jinshuo Cui\textsuperscript{1}} \quad \mbox{Junbo Cui\textsuperscript{6}} \quad \mbox{Kezhao Zhao\textsuperscript{6}} \quad \mbox{Longxi Gao\textsuperscript{1}} \quad \mbox{Mengwei Xu\textsuperscript{*,1,5}} \quad \mbox{Rongjie Yi\textsuperscript{1,5}} \quad \mbox{Ruixin Liu\textsuperscript{3}} \quad \mbox{Shangguang Wang\textsuperscript{1}} \quad \mbox{Tam Sikyuen\textsuperscript{4,6}} \quad \mbox{Tianyue Zhang\textsuperscript{1}}
        \mbox{Weikai Xie\textsuperscript{1}} \quad \mbox{Xuanzhe Liu\textsuperscript{*,3}} \quad \mbox{Yingying Qin\textsuperscript{1}} \quad \mbox{Yiwen Lu\textsuperscript{1}} \quad \mbox{Yuan Yao\textsuperscript{*,4,6}} \quad \mbox{Yuezhi Zu\textsuperscript{4}} \quad \mbox{Yunhan Guo\textsuperscript{1}} \quad \mbox{Yuxin Zheng\textsuperscript{6}} \quad \mbox{Ziqi Guo\textsuperscript{1}}\par}
    \vspace{3pt}
    {\sffamily\fontsize{7.4}{9}\selectfont
        \textsuperscript{\textdagger}Project Lead \quad \textsuperscript{*}Corresponding authors\par
        \vspace{1pt}
        \textsuperscript{1}Beijing University of Posts and Telecommunications \quad \textsuperscript{2}Nanjing University \quad \textsuperscript{3}Peking University\par
        \textsuperscript{4}Tsinghua University \quad \textsuperscript{5}MingTi Technology \quad \textsuperscript{6}ModelBest\par}
    \vspace{7pt}
}
}
\end{adjustwidth}
\egroup

{\abscontent}
\vspace{3pt}

\begin{figure}[H]
    \centering
    \captionsetup{font=footnotesize,skip=2pt}
    \includegraphics[width=\linewidth]{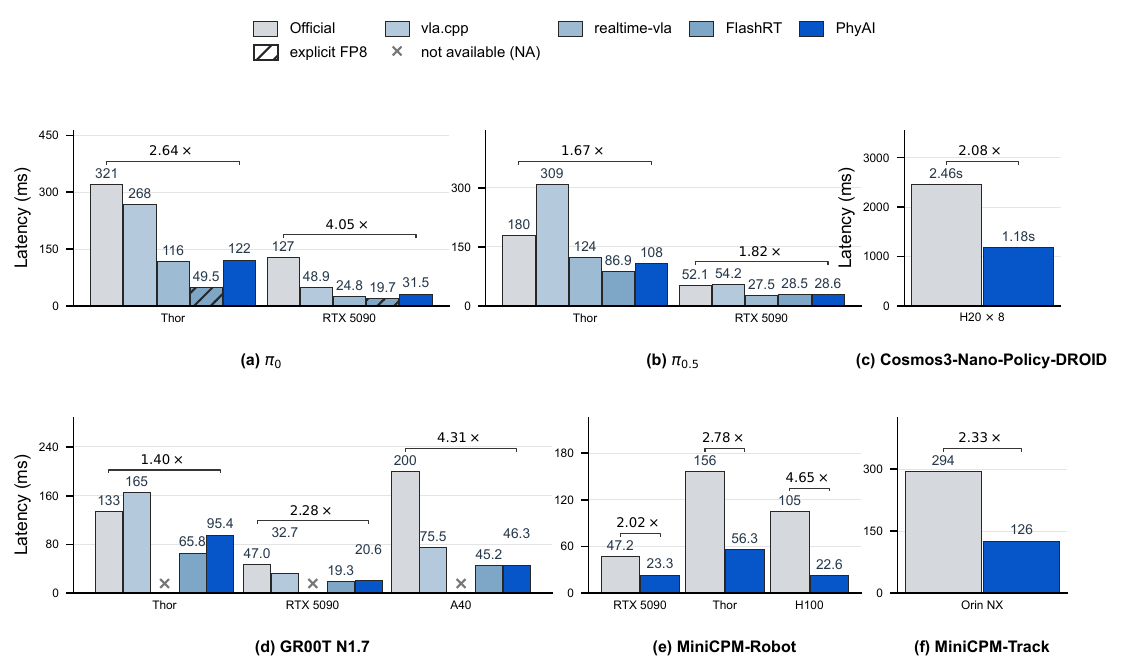}
    \caption{Single-request latency across six embodied policies. \textbf{(a)}, $\pi_0$. \textbf{(b)}, $\pi_{0.5}$. \textbf{(c)}, Cosmos3-Nano-Policy-DROID. \textbf{(d)}, GR00T N1.7. \textbf{(e)}, MiniCPM-Robot. \textbf{(f)}, MiniCPM-Track. Brackets give PhyAI speedup over the official path on matched hardware; lower is better. Hatched bars mark explicit FP8 results, and crosses mark unavailable measurements. MiniCPM-Robot uses LIBERO with two $224\times224$ views; Cosmos3 uses eight H20 GPUs with CFG=2 and TP=4; MiniCPM-Track is measured on Orin NX.}
    \label{fig:speedup-summary}
\end{figure}

\thispagestyle{firststyle}
\clearpage
}

\input{01-intro}
\input{02-background}
\input{03-method}
\input{04-results}
\input{05-conclusion-discussion}

\clearpage
\begingroup
\normalsize
\setlength{\bibsep}{3pt}
\bibliography{main}
\endgroup

\clearpage
\input{06-appendix}
\end{document}

%% file: 00-abs.tex
Physical AI policies require inference throughout their lifecycle, including model evaluation, cloud reinforcement learning (RL) rollout, edge GPU serving, and onboard deployment. Although these settings use the same checkpoint and action semantics, they often rely on separate inference programs. To provide one inference path across these settings, we build \textbf{PhyAI}, an \textbf{Phy}sical \textbf{AI} inference engine with a unified runtime. PhyAI keeps architecture-specific conditioning, solver, cache, and output logic in model adapters while sharing graph execution, kernels, memory management, and parallel services. The same codebase runs vision-language-action (VLA) models and world-action models (WAMs) on single or multiple GPUs across onboard, edge, and cloud deployments. We used the same adapter interface to add MiniCPM-Robot on the day of its release. PhyAI achieves $1.40\times$ to $4.65\times$ speedups over the official implementations of $\pi_0$, $\pi_{0.5}$, GR00T N1.7, and MiniCPM-Robot. On Cosmos3-Nano-Policy-DROID, it reduces latency from $2.46$ to $1.18$~s on eight H20 GPUs with CFG=2 and TP=4, a $2.08\times$ speedup. Specialized runtimes remain faster in several configurations, so our goal is one runtime with competitive latency rather than the fastest result in every case. The detailed profiles also show why these models need different execution policies. On a Hopper-series GPU at batch size one, the $\pi_{0.5}$ action expert accounts for $8.8\%$ of estimated FLOPs but $57.2\%$ of profiled latency. At batch size 32, its share falls to $13.5\%$, throughput reaches about 100 samples/s, and the full batch takes about 320ms. Cosmos3 remains generation dominated and gains only $14.3\%$ throughput as batch size increases from 1 to 16. We further introduce the \emph{control-time Roofline}, which distinguishes inference-bound from environment-bound control. The measured $\pi_{0.5}$ points on four LIBERO suites are environment bound, while Cosmos3 remains inference bound. Code and benchmark harnesses are available at \href{https://github.com/mingti-org/phyai}{github.com/mingti-org/phyai}.

%% file: 01-intro.tex
% note(chenghua): reviewed and modified in 2026-07-26. Almost Done.
\section{Introduction}
\label{sec:introduction}
\suppressfloats[t]

Physical AI policies now span vision-language-action (VLA) models such as RT-1, RT-2, $\pi_0$, $\pi_{0.5}$, and GR00T, as well as world-action models (WAMs) such as Cosmos3 and DreamZero \citep{rt1,rt2,pi_0,pi_05,gr00t,cosmos3,dreamzero,wamsurvey,motionwam}. The same policy may be used for offline evaluation, cloud reinforcement-learning (RL) rollout, shared edge serving, and onboard deployment. These settings should preserve the same preprocessing, solver, and action semantics, but they often rely on separate inference programs. Each path must then be optimized and validated independently, and behavior established in one setting does not automatically carry to another.

A common inference path is difficult to build because embodied models differ in both execution structure and deployment requirements. An embodied request includes observation capture, preprocessing, model inference, output conversion, and action execution. A $\pi$-family VLA runs a vision-language backbone followed by an iterative action expert, whereas a Cosmos3 WAM evolves video and action states together. Onboard execution emphasizes single-request latency, shared edge serving must balance latency against throughput, and cloud rollout uses larger batches and distributed execution. These workloads require different scheduling and parallel strategies even when they share the same policy interface.

Existing systems cover parts of this problem. FlashRT, vla.cpp, realtime-vla, and Embodied.cpp optimize selected models or devices \citep{flashrt,vlacpp,realtimevla,embodiedcpp}. LeRobot covers data collection, training, evaluation, and a generalized asynchronous policy-inference stack that can decouple remote action prediction from robot control. It does not, however, provide the same kernel-level, multi-architecture execution engine targeted by PhyAI \citep{lerobot}. vLLM and SGLang offer mature token-oriented LLM serving \citep{vllm,sglang}, but their execution model does not directly represent reusable multimodal conditions, iterative action solvers, evolving world latents, or classifier-free-guidance branches. The missing component is an inference runtime that preserves model semantics while allowing each architecture and deployment setting to choose an appropriate execution policy.

\noindent\textbf{PhyAI runtime.}
We build PhyAI, a unified inference runtime for embodied policies. PhyAI separates model-specific logic from shared execution services. Model adapters retain conditioning, solver state, reusable tensors, scheduling, and action conversion. The runtime provides graph replay, optimized kernels, memory and cache management, quantization paths, communication, and parallel execution. This separation allows the $\pi_{0.5}$ path to reuse its single-GPU runner for data-parallel execution, while Cosmos3 combines tensor parallelism with classifier-free-guidance parallelism.

The same codebase currently runs $\pi_0$, $\pi_{0.5}$, GR00T N1.7, Cosmos3-Nano-Policy, Cosmos3-Edge, and MiniCPM-Robot on devices ranging from Jetson Thor to RTX~5090, A40, H100, and multi-GPU H20 and A100 servers. Local inference, distributed execution, benchmarks, and rollout workers call the same model path. For cloud RL, PhyAI integrates at the RLinf inference-backend boundary \citep{rlinf}. The adapter interface also allowed us to add MiniCPM-Robot on the day of its release without introducing a separate runtime.

\noindent\textbf{Control-time Roofline.}
Lower inference latency does not always produce a proportional improvement in control rate. In a sequential loop, model inference and environment execution occur one after the other. Real-time chunking instead generates the next action chunk while the current chunk is being executed \citep{rtc}. In this setting, the relevant deadline is the next action handoff. Inference that finishes before the handoff is hidden within the current execution window, while inference that finishes later remains on the critical path.

We introduce the \emph{control-time Roofline} to describe this boundary. Let $L_{\mathrm{inference}}$ denote the inference time and $L_{\mathrm{env}}$ the time required to execute the current action and advance the environment. Under ideal overlap, the slower stage determines the control period:
\[
L_{\mathrm{overlap}}
=
\max\!\left(L_{\mathrm{inference}},L_{\mathrm{env}}\right).
\]
The Roofline compares the inference rate with the environment execution rate. When inference is slower, the system is inference bound and lower model latency directly improves control. When environment execution is slower, the system is environment bound and further acceleration creates timing margin rather than a higher ideal control rate. That margin can absorb runtime variation or support more policies per GPU, a less expensive device, or a larger model.

\noindent\textbf{Evaluation.}
We first measure single-request latency across the available models and devices. PhyAI is faster than the corresponding official path in all 11 measured pairs, with speedups from $1.40\times$ to $4.65\times$. On Cosmos3-Nano-Policy-DROID, PhyAI reduces latency from $2.46$ to $1.18$~s on eight H20 GPUs with CFG=2 and TP=4, a $2.08\times$ speedup. Specialized runtimes remain faster in several configurations, and the available comparisons are not fully precision matched. We therefore use the official implementations as the common reference rather than claiming the fastest result in every configuration.

The phase and batch profiles explain why these models require different execution policies. On a Hopper-series GPU at batch size one, the $\pi_{0.5}$ action expert accounts for only $8.8\%$ of estimated FLOPs but $57.2\%$ of profiled latency. At batch size 32, its latency share falls to $13.5\%$, throughput reaches about 100 samples/s, and the full synchronized batch takes about 320~ms. On RTX~5090, GR00T reaches $92.2\%$ of its batch-size-32 throughput by batch size eight, while its main bottleneck shifts from Action Head at batch size one to Backbone from batch size four onward. Cosmos3 is already dominated by its generation phase and gains only $14.3\%$ throughput when the batch size increases from 1 to 16. In the control-time analysis, the measured $\pi_{0.5}$ points on four LIBERO suites are environment bound, while the Cosmos3 generation path remains inference bound.

We further simulated an integrated RL rollout workload with PhyAI as its inference backend. The setup used eight A100 GPUs, a batch size of 40, and 41 policy-inference calls per RL step. Inference accounted for $53.1\%$ of the baseline step time and $36.2\%$ with PhyAI, a decrease of 16.9 percentage points, or $31.8\%$ relative. With the non-inference time held fixed in the simulation, this change corresponds to a $26.5\%$ reduction in rollout-step latency and about $1.36\times$ higher training throughput.

We make the following contributions:
\begin{itemize}
    \item \textbf{Unified inference runtime.} PhyAI provides one model path for VLA and WAM inference across onboard, edge, and cloud execution. Model adapters preserve architecture-specific semantics while shared services provide graph execution, kernels, memory management, quantization, communication, and parallelism.
    \item \textbf{Control-time Roofline.} The analysis relates inference latency to environment execution time and distinguishes inference-bound from environment-bound control under an action-handoff deadline.
    \item \textbf{System evaluation.} We evaluate single-request latency, phase behavior, static-batch scaling, multi-GPU execution, and a simulated RL rollout workload across VLA and WAM models.
\end{itemize}

The code and benchmark harnesses are available at
\href{https://github.com/mingti-org/phyai}{github.com/mingti-org/phyai}.

%% file: 02-background.tex
\section{Related Work}
\label{sec:related-work}
\suppressfloats[t]

% TODO(chenghua): reviewed and modified in 2026-07-26. We need check each citation in google scholar.

\subsection{Embodied Model Architectures}
\label{sec:embodied-paradigms}

% TODO(chenghua): reviewed and modified in 2026-07-26. Longxi Gao write this section. while I think it still need to be polished.

Existing embodied policies fall into three inference families. Autoregressive action-token VLAs, including RT-2\citep{rt2} and $\pi_0$-FAST\citep{pi_0_fast}, tokenize continuous actions and decode them causally with a VLM. This design reuses language-model decoding, but tokenization can lose fine-grained action information, and serial generation increases latency with sequence length. These costs motivate continuous-action and world-action policies.

Continuous-action VLAs generate action chunks by flow matching or diffusion. Table~\ref{tab:embodied-architecture-taxonomy} groups them by how vision-language features condition action generation. Prefix-coupled models such as the $\pi$ family\citep{pi_0,pi_05,pi_07} feed reusable prefix states to a compact action expert; feature-handoff models such as GR00T\citep{gr00t} pass fixed VLM features to a separate iterative head; single-stack flow models such as RynnBrain-VLA\citep{rynnbrain-1,rynnbrain-1_1} use one full-depth transformer for conditioning and denoising. These structures expose distinct bottlenecks: small-operator overhead, repeated access to fixed features, or repeated execution of a deep backbone.

World-action models couple action prediction with future visual or latent dynamics. Models such as Cosmos3\citep{cosmos3} and DreamZero\citep{dreamzero} maintain both world and action states during inference, increasing the size of the evolving state and potentially adding guidance branches or cross-device communication. Other WAMs use future prediction only during training and disable world generation at deployment\citep{fastwam,gigaworld-policy-0,gigaworld-policy-0_5}. RynnVLA-001 is an example of this action-only deployment pattern: video-generation pretraining provides auxiliary supervision, while inference produces an action embedding that ActionVAE decodes into an action sequence\citep{rynnvla-001}. These families differ in reusable state, evolving generation state, operator shapes, and parallel structure, and cannot be reduced to a single LLM-style prefill--decode pattern.

\begin{table}[htbp]
    \centering
    \footnotesize
    \setlength{\tabcolsep}{3.8pt}
    \renewcommand{\arraystretch}{1.2}
    \begin{tabularx}{0.998\linewidth}{@{}L{0.18\linewidth}L{0.19\linewidth}Y L{0.28\linewidth}@{}}
        \toprule
        \textbf{Model family} & \textbf{Architecture} & \textbf{Inference structure} & \textbf{Representative works} \\
        \midrule

        Autoregressive VLA
        & Action-token decoder
        & Continuous actions are tokenized and decoded causally within the VLM sequence.
        & RT-2\citep{rt2}; OpenVLA\citep{openvla}; $\pi_0$-FAST\citep{pi_0_fast}; Galaxea G0.5 \citep{galaxea-g05} \\

        \midrule

        \multirow{3}{=}{Continuous-action VLA}
        & Prefix-coupled expert
        & A reusable VLM prefix or KV state conditions a smaller iterative action expert.
        & $\pi$ family\citep{pi_0,pi_05,pi_07}; LingBot-VLA\citep{lingbot-vla}; Hy-Embodied-0.5-VLA\citep{hy-embodied}; Xiaomi-Robotics-0/1\citep{xiaomi-robotics-0,xiaomi-robotics-1}; GigaBrain-0\citep{gigabrain-0} \\

        & Feature-handoff head
        & A VLM runs once and passes hidden features to a separate iterative action head.
        & GR00T N1.x\citep{gr00t}; CogACT\citep{cogact}; ABot-M0\citep{abot-m0} \\

        & Single-stack flow
        & Vision, language, state, and noisy actions share one denoising transformer stack.
        & RynnBrain-VLA 1.0/1.1\citep{rynnbrain-1,rynnbrain-1_1} \\

        \midrule

        World-action model
        & Joint world-action generation
        & Future world states and actions are predicted jointly or through tightly coupled paths.
        & DreamZero\citep{dreamzero}; LingBot-VA\citep{lingbotva-2}; Cosmos3\citep{cosmos3}; RynnVLA-002\citep{rynnvla-002}; GigaWorld-Policy 0/0.5\citep{gigaworld-policy-0,gigaworld-policy-0_5}; ABot-M0.5\citep{abot-m0_5} \\

        \bottomrule
    \end{tabularx}
    \caption{An inference-oriented taxonomy of embodied policy architectures. Continuous-action VLAs are further divided by how vision-language representations are coupled to iterative action generation.}
    \label{tab:embodied-architecture-taxonomy}
\end{table}
\FloatBarrier

\subsection{Training, Inference, and RL Infrastructure}
% note(wangchenghua): This section is already polished, do not modify it.
% note(wangchenghua): This section is already polished, do not modify it.
% note(wangchenghua): This section is already polished, do not modify it.

Embodied model infrastructure is split across the model lifecycle. LeRobot provides checkpoint loading, processors, training code, reference behavior, and a generalized asynchronous policy-inference stack, while StarVLA packages modular training recipes and benchmark interfaces \citep{lerobot,starvla}. RLinf coordinates distributed embodied RL and delegates policy execution to an inference backend \citep{rlinf}. These systems provide lifecycle, orchestration, and policy-inference interfaces, but do not expose the same kernel-level execution scope as PhyAI.

Existing inference runtimes cover distinct parts of the deployment stack. FlashRT optimizes latency-sensitive VLA and WAM execution, vla.cpp packages VLA policies in a portable C++ runtime, and realtime-vla provides model-specific Triton implementations for $\pi_0$, $\pi_{0.5}$, and DM0 \citep{flashrt,flashrt-project,vlacpp,realtimevla}. \Cref{tab:runtime-scope} includes only capabilities documented by each project at the time of writing. We reviewed the linked public repositories on 24 July 2026.\footnote{Official projects: \href{https://github.com/mingti-org/phyai}{PhyAI}, \href{https://github.com/flashrt-project/FlashRT}{FlashRT}, \href{https://github.com/VinRobotics/vla.cpp}{vla.cpp}, and \href{https://github.com/Dexmal/realtime-vla}{realtime-vla}. The FlashRT INT8 entry refers to the publicly documented Jetson AGX Orin path in \href{https://github.com/flashrt-project/FlashRT/blob/main/docs/deployment_orin.md}{deployment\_orin.md}.}

\begin{table}[htbp]
    \centering
    \footnotesize
    \setlength{\tabcolsep}{2.6pt}
    \renewcommand{\arraystretch}{1.22}
    \begin{tabularx}{\linewidth}{@{}L{0.17\linewidth}*{4}{>{\centering\arraybackslash}X}L{0.14\linewidth}Y@{}}
        \toprule
        \textbf{Framework} & \textbf{VLA} & \textbf{WAM} & \makecell{\textbf{NVFP4/}\\\textbf{FP8}} & \makecell{\textbf{INT4/}\\\textbf{INT8}} & \textbf{Deployment} & \makecell[l]{\textbf{Parallelism}} \\
        \midrule
        PhyAI (ours) & \ding{51} & \ding{51} & \ding{51} & \ding{51} & Edge + Cloud & DP/TP/CFG \\
        FlashRT & \ding{51} & \ding{51} & \ding{51} & \ding{51} & Edge + Cloud & -- \\
        vla.cpp & \ding{51} & -- & -- & -- & Edge & -- \\
        realtime-vla & \ding{51} & -- & -- & -- & Edge & -- \\
        \bottomrule
    \end{tabularx}
    \caption{Publicly documented capabilities of inference runtimes. Checkmarks indicate documented support; dashes indicate that no public claim was found. FlashRT INT4/INT8 refers to the community Jetson AGX Orin INT8 path documented by the project.}
    \label{tab:runtime-scope}
\end{table}
\FloatBarrier

LLM systems commonly reuse the same inference implementation for evaluation, RL rollouts, and deployment. Physical AI often requires separate integration for each setting. PhyAI aims to provide the same workflow through a common inference backend, while existing systems continue to handle training and RL orchestration.

\subsection{Acceleration Methods}
% note(wangchenghua): This section is already polished, do not modify it.
% note(wangchenghua): This section is already polished, do not modify it.
% note(wangchenghua): This section is already polished, do not modify it.

Recent work speeds up embodied models by reducing computation, reusing intermediate states, or overlapping inference with action execution. \Cref{tab:embodied-acceleration} groups these VLA and WAM specific methods by the main source of their speedup.

\begin{table}[htbp]
    \centering
    \footnotesize
    \setlength{\tabcolsep}{3.8pt}
    \renewcommand{\arraystretch}{1.2}
    \begin{tabularx}{0.998\linewidth}{@{}L{0.16\linewidth}L{0.18\linewidth}L{0.08\linewidth}Y@{}}
        \toprule
        \textbf{Method family} & \textbf{Method} & \textbf{Model} & \textbf{Primary mechanism} \\
        \midrule
        \multirow{2}{*}{Quantization}
        & QuantVLA \citep{quantvla} & VLA & Mixed precision calibrated for the language backbone and DiT action head \\
        & ActQuant \citep{actquant} & VLA & Sub 4bits precision assigned by each tensor's effect on action prediction \\
        \midrule
        \multirow{2}{*}{Sparsification}
        & EfficientVLA \citep{efficientvla} & VLA & Language layer pruning, task aware visual token selection, and DiT feature reuse \\
        & VLA-Pruner \citep{vlapruner} & VLA & Visual token pruning based on semantic and action relevance \\
        \midrule
        \multirow{4}{*}{State reuse}
        & OxyGen \citep{oxygen} & VLA & Shared prefix KV across tasks and continuous batching across frames \\
        & VLA-Cache \citep{vlacache} & VLA & KV reuse for stable visual tokens across observations \\
        & ActionCache \citep{actioncache} & VLA & Retrieved intermediate action states initialize flow generation \\
        & \mbox{\ensuremath{\mathrm{C}^{3}\mathrm{ache}}} \citep{c3ache} & WAM & Denoising residual reuse between action chunks at the same timestep \\
        \midrule
        \multirow{2}{=}{Generation reduction}
        & SnapFlow \citep{snapflow} & VLA & Distillation of multi step flow matching into one generation step \\
        & Fast-WAM \citep{fastwam} & WAM & Future video co-training with video generation removed at inference \\
        \midrule
        Control scheduling & RTC \citep{rtc} & VLA & Generate the next action chunk while the current chunk executes \\
        \bottomrule
    \end{tabularx}
    \caption{VLA and WAM specific acceleration methods grouped by their main mechanism. Reported speedups are not directly comparable because the methods use different models, hardware, action horizons, and task suites.}
    \label{tab:embodied-acceleration}
\end{table}

QuantVLA uses VLA specific calibration to assign precision across the language backbone and DiT action head. ActQuant uses each tensor's influence on predicted actions to allocate sub 4bits precision \citep{quantvla,actquant}. EfficientVLA reduces computation through language layer pruning, task-aware visual token selection, and feature reuse in the DiT action head \citep{efficientvla}. VLA-Pruner scores visual tokens using both semantic and action relevance, preserving tokens that matter to control even when they contribute little to language processing \citep{vlapruner}. Because these methods change the model computation, their speedups may come with changes in task performance.

State reuse can be exact or approximate. Conditions and embeddings that stay fixed across denoising steps can be cached without changing the result. OxyGen shares prefix KV across language and action tasks and batches work across frames \citep{oxygen}. VLA-Cache reuses KV for visually stable tokens while recomputing task relevant tokens \citep{vlacache}. ActionCache retrieves intermediate action states from similar contexts to initialize a new flow trajectory \citep{actioncache}. \mbox{\ensuremath{\mathrm{C}^{3}\mathrm{ache}}} reuses WAM residuals between action chunks at the same denoising step \citep{c3ache}. OxyGen preserves exact computation, whereas the other three methods reuse state across changed inputs and may alter the output.

SnapFlow distills a multi-step flow policy into a single generation step \citep{snapflow}. Fast-WAM keeps future video prediction during training but removes it from inference \citep{fastwam}. Both reduce latency by changing the generation procedure, which may also change task behavior.

RTC changes the control schedule rather than the model computation. It generates the next action chunk while the robot executes the current one \citep{rtc}. This overlap reduces idle time, but the next chunk depends on an older observation. Quantization, state reuse, shorter generation, and RTC can be combined because they act on different parts of the inference and control pipeline.

\FloatBarrier
\section{Embodied Task Formulation}
\label{sec:background}
\suppressfloats[t]

\subsection{From Observation to Action}
% note(wangchenghua): This section is already polished, do not modify it.
% note(wangchenghua): This section is already polished, do not modify it.
% note(wangchenghua): This section is already polished, do not modify it.

At control step $t$, the policy receives recent camera observations $o_t$, a language instruction $\ell$, and the robot state $s_t$. It maps these inputs to the next $H$ actions:

\begin{equation}
    x_t=(o_t,\ell,s_t),
    \qquad
    a_{t:t+H-1}=\pi_\theta(x_t).
\end{equation}

For diffusion and flow policies, $\pi_\theta$ may refine an action latent over several steps. All latency terms below refer to one policy request that returns one $H$-action chunk. The observation-to-action interval starts when sensor capture begins and ends when the low-level controller accepts the returned chunk.

$L_{\mathrm{observe}}$ covers sensor exposure, readout, synchronization, and preparation of model inputs. $L_{\mathrm{transfer}}$ is the total transport time between the control host and the inference device, including serialization, device copies, and both directions of remote communication when applicable. Once the request reaches the runtime, $L_{\mathrm{queue}}$ measures the wait for batching and scheduling. $L_{\mathrm{inference}}$ starts when the model runner begins execution and ends when the complete action chunk is ready; it includes conditioning, iterative generation, and model-side output decoding. Finally, $L_{\mathrm{actuate}}$ covers controller-side conversion and handoff of the returned actions. With these boundaries, the full path is

\begin{equation}
    L_{\mathrm{critical}}
    =L_{\mathrm{observe}}
    +L_{\mathrm{transfer}}
    +L_{\mathrm{queue}}
    +L_{\mathrm{inference}}
    +L_{\mathrm{actuate}}.
    \label{eq:critical-latency}
\end{equation}

$L_{\mathrm{critical}}$ ends at controller handoff and does not include physical execution of the returned actions. Onboard inference removes the network component of $L_{\mathrm{transfer}}$, but local sensor and device copies may remain. A shared edge or cloud service adds network transport and queueing. Observation and actuation remain even when inference becomes faster.

\subsection{Control-Time Roofline}
% note(wangchenghua): This section is already polished, do not modify it.
% note(wangchenghua): This section is already polished, do not modify it.
% note(wangchenghua): This section is already polished, do not modify it.

An action chunk is useful only if it arrives before its handoff deadline. We group the time spent outside the model as

\begin{equation}
    L_{\mathrm{noninf}}
    =L_{\mathrm{observe}}+L_{\mathrm{transfer}}+L_{\mathrm{queue}}+L_{\mathrm{actuate}}.
\end{equation}

Let $B_{\mathrm{ctrl}}$ be the allowed time from the start of observation capture to controller handoff. The remaining margin is

\begin{equation}
    M_{\mathrm{ctrl}}
    =B_{\mathrm{ctrl}}-\left(L_{\mathrm{inference}}+L_{\mathrm{noninf}}\right).
    \label{eq:control-margin}
\end{equation}

A positive margin means that the action is ready before the deadline. A deployment should measure the distribution of this margin, not only its mean. For example, it may require $P(M_{\mathrm{ctrl}}\geq 0)\geq 1-\epsilon$.

The control-time Roofline relates request latency to the time available between policy decisions. Let $L_{\mathrm{env}}$ be the interval from the handoff of the current action chunk until the next policy decision is due. In simulation, it covers the actions actually applied and the corresponding environment steps. A controller may replan before using all $H$ predicted actions, so this interval need not equal the full predicted horizon.

For a deployed controller, sequential execution and ideal overlap give

\begin{equation}
    L_{\mathrm{seq}}^{\mathrm{full}}=L_{\mathrm{critical}}+L_{\mathrm{env}},
    \qquad
    L_{\mathrm{overlap}}^{\mathrm{full}}=\max(L_{\mathrm{critical}},L_{\mathrm{env}}).
\end{equation}

The LIBERO timing records contain per-decision model inference and environment time, but not observation, transfer, queueing, or actuator handoff. For the plot, we therefore use the inference-only periods

\begin{equation}
    L_{\mathrm{seq}}=L_{\mathrm{inference}}+L_{\mathrm{env}},
    \qquad
    L_{\mathrm{overlap}}=\max(L_{\mathrm{inference}},L_{\mathrm{env}}).
\end{equation}

Here, $L_{\mathrm{control}}$ is either $L_{\mathrm{seq}}$ or $L_{\mathrm{overlap}}$, depending on the schedule. Ideal overlap gives a lower bound on the control period and, equivalently, an upper bound on the control rate. We normalize by $L_{\mathrm{env}}$ so that the LIBERO suites share the same axes:

\begin{equation}
    X=\frac{L_{\mathrm{env}}}{L_{\mathrm{inference}}},
    \qquad
    Y=\frac{L_{\mathrm{env}}}{L_{\mathrm{control}}}.
\end{equation}

The horizontal axis, $X$, is the inference rate relative to the environment execution rate. For example, $X=2$ means that inference is twice as fast as environment execution. Values below one are inference bound; values above one are environment bound. The vertical axis, $Y$, is the control rate relative to the environment execution limit. For example, $Y=0.8$ means that the loop reaches 80\% of that limit.

Substituting the two control periods gives the two curves in the figure:

\begin{equation}
    Y_{\mathrm{seq}}=\frac{X}{1+X},
    \qquad
    Y_{\mathrm{roof}}=\min(X,1).
\end{equation}

The measurements follow the first curve because their reported control period is the sum of inference and environment time. The second curve is the ideal overlap roof. Once $X$ exceeds one, environment execution sets this inference-only roof. Unmeasured transport, queueing, and controller overhead can still place a deployed loop on the critical path.

\begin{figure}[tbp]
    \centering
    \captionsetup[subfigure]{font={small,bf,color=black},labelfont={small,bf,color=black},labelsep=space,justification=centering,singlelinecheck=true,skip=1pt}
    \begin{subfigure}[t]{0.82\linewidth}
        \centering
        \includegraphics[width=\linewidth]{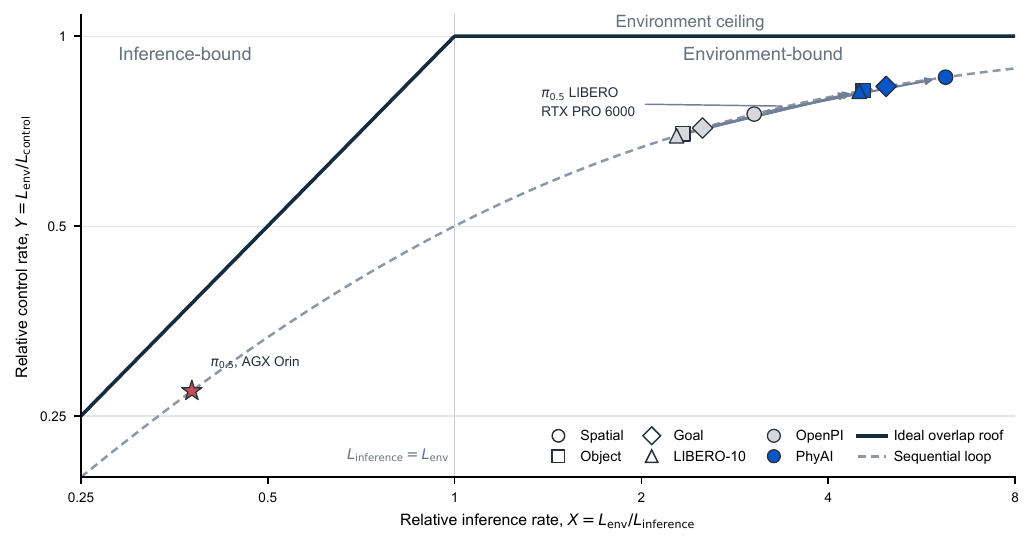}
        \caption{Control-time Roofline for four LIBERO suites.}
        \label{fig:libero-control-roofline}
    \end{subfigure}
    \vspace{2pt}

    \begin{subfigure}[t]{0.80\linewidth}
        \centering
        \includegraphics[width=\linewidth]{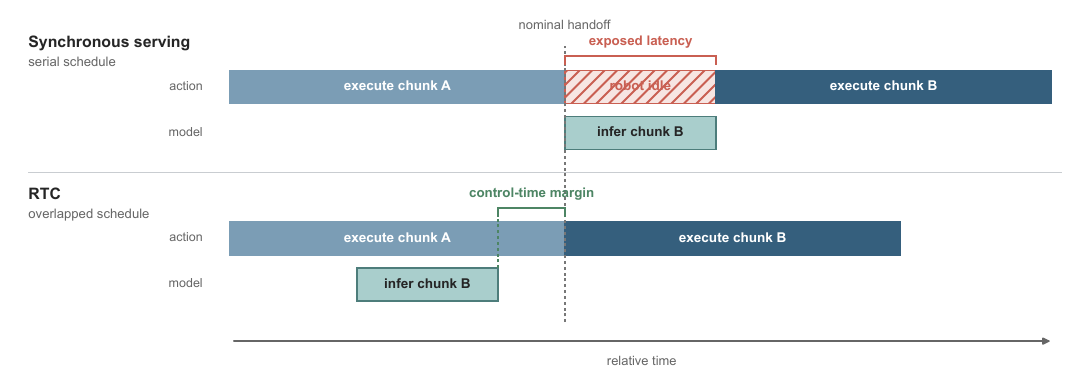}
        \caption{Timing of synchronous serving and RTC.}
        \label{fig:rtc-window}
    \end{subfigure}
    \caption{Control-time Roofline and RTC schedules. \textbf{(a)}, Mean control rates for OpenPI and PhyAI on four LIBERO suites. \textbf{(b)}, Schematic timing of synchronous serving and RTC across two action chunks.}
    \label{fig:control-time-analysis}
\end{figure}

RTC changes the schedule by preparing the next action chunk during the current execution window. In the inference-only view shown in \Cref{fig:rtc-window}, it can hide model latency that finishes before handoff. A deployed service must also complete transfer, queueing, and controller handoff within that window. Executing more actions before replanning creates more time for the request, but the actions come from an older observation. Evaluation must therefore vary the executed action horizon together with latency and task outcome.

\subsection{Edge and Edge-Cloud Deployment}
% note(wangchenghua): This section is already polished, do not modify it.
% note(wangchenghua): This section is already polished, do not modify it.
% note(wangchenghua): This section is already polished, do not modify it.

Onboard execution connects directly to cameras and control hardware, so inference latency can translate into a fresher action until another component becomes the limit. Lower latency is still useful after that point, but its role changes. The recovered time may absorb p99 jitter, run a validation check, or allow a smaller and cheaper device to meet the same deadline.

Running the model on a shared edge server(e.g.: RTX Pro 6000) can reduce per-robot hardware cost and improve GPU utilization. It also adds network and queueing delay, so the action must still arrive before its deadline. When requests are batched, a robot waits for the whole batch to finish. The average time per sample does not show how long the robot waits. RTC gives inference more time by overlapping it with action execution, but it does not reduce network or queueing delay.

\subsection{Cloud Batching}
% note(wangchenghua): This section is already polished, do not modify it.
% note(wangchenghua): This section is already polished, do not modify it.
% note(wangchenghua): This section is already polished, do not modify it.

Cloud RL rollout and shared model serving must track both batch latency and throughput. For a synchronized batch of $B$ requests, let $T_{\mathrm{sync}}(B)$ be the time from launch until all outputs are ready. We report

\begin{equation}
    Q(B)=\frac{B}{T_{\mathrm{sync}}(B)},
    \qquad
    T_{\mathrm{amort}}(B)=\frac{T_{\mathrm{sync}}(B)}{B}=\frac{1}{Q(B)}.
    \label{eq:batch-throughput}
\end{equation}

$Q(B)$ is the number of completed requests per unit time. $T_{\mathrm{amort}}(B)$ divides the batch time across the $B$ requests. Every request still waits $T_{\mathrm{sync}}(B)$, so $T_{\mathrm{amort}}(B)$ is not request latency.

Faster inference can raise $Q(B)$ and shorten $T_{\mathrm{sync}}(B)$. This may reduce the number of GPUs needed for a target load. Full request latency also includes network transfer and batch formation, which should be measured separately.

A faster runtime also leaves more latency budget for the model and controller. The system can use this budget for a larger model, higher visual resolution, more solver steps, or a shorter action horizon. The phase profile shows which resource limits each option: compute, memory traffic, or kernel launch overhead.

\FloatBarrier

%% file: 03-method.tex
\section{PhyAI Architecture}
\label{sec:methodology}
\suppressfloats[t]

PhyAI is organized around a boundary between model semantics and execution policy. A model adapter owns the operations that define the policy, including preprocessing, phase order, solver updates, cache validity, and action conversion. The runtime manages scheduling, memory, parallel execution, and operator selection. This separation keeps the model path unchanged when its batch size, device placement, or hardware backend changes.

The boundary also reflects the control and deployment constraints described in the previous section. Different models expose different opportunities for reuse and parallelism, while onboard, edge, and cloud workloads place different demands on latency and throughput. PhyAI therefore shares execution services across models while leaving each adapter's generation procedure intact. Reused state carries an explicit validity scope, and execution choices are made for the model phase, tensor shape, and target device.

\subsection{Overview}

\begin{figure}[tbp]
    \centering
    \includegraphics[width=0.98\linewidth]{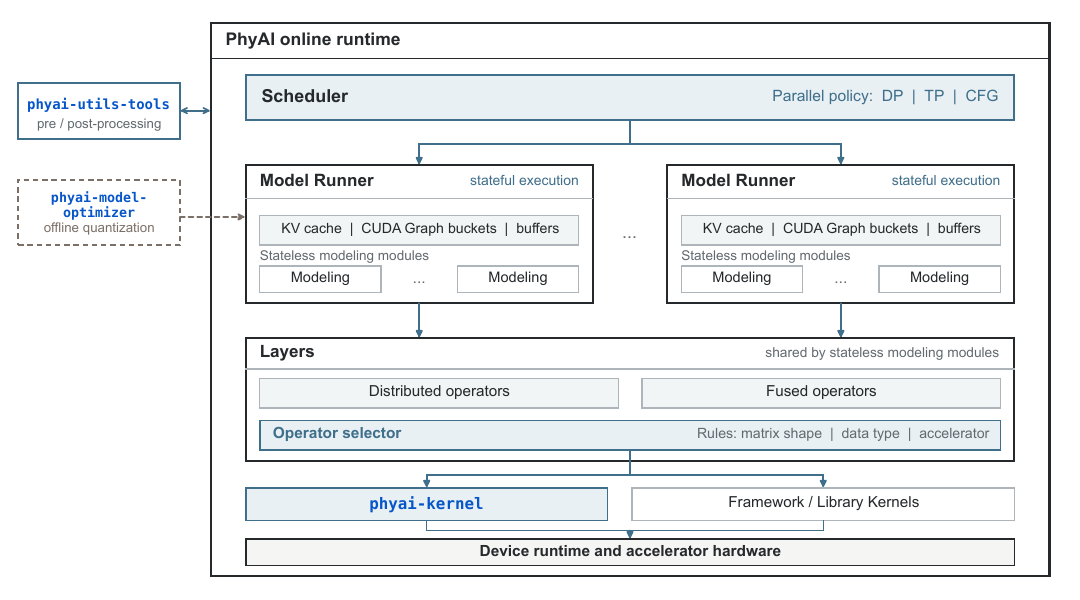}
    \caption{Layered PhyAI architecture and component ownership.}
    \label{fig:architecture}
\end{figure}

\Cref{fig:architecture} shows the ownership hierarchy of the PhyAI runtime. A scheduler manages multiple model runners, assigns requests to them, and configures data parallelism (DP), tensor parallelism (TP), and classifier-free-guidance (CFG) parallelism. Device groups and work partitioning are decided at this level. The scheduler determines where a request runs, but it does not hold model tensors or implement model computation.

Each model runner owns the mutable state of its requests and can invoke several modeling modules along one execution path. It manages intermediate buffers, KV caches, CUDA Graph buckets, and their memory lifetimes. The runner also advances the solver or generation loop and assembles the resulting action chunk. Modeling modules are stateless. They implement architecture-specific computation, such as multimodal prefix encoding, action experts, and coupled video-action generation, without retaining request state between calls.

Modeling modules express their computation through the Layers interface. Layers provide fused and distributed operators and encapsulate the operator selector. For each operation, the selector applies rules over the matrix shape, data type, accelerator model, and execution configuration, then dispatches a suitable kernel from \textcolor{phyaiblue}{\texttt{phyai-kernel}}, the framework backend, or an external library. Outside the online path, \textcolor{phyaiblue}{\texttt{phyai-utils-tools}} handles preprocessing and post-processing, while \textcolor{phyaiblue}{\texttt{phyai-model-optimizer}} quantizes models before deployment.

\subsection{Optimization Methods}

Once the scheduler assigns a request to a model runner, we optimize the work performed along that execution path. The runner removes repeated setup and reuses state whose validity spans multiple model evaluations. Below the stateless modeling interface, Layers fuse operators and select kernels for the observed tensor shapes and hardware, while the offline optimizer prepares low-precision model artifacts. These methods leave the model implementation unchanged; the next subsection describes how the scheduler distributes the resulting path across devices.

\paragraph{Kernel fusion.}
For small batches, launch overhead and intermediate tensor traffic can dominate short operators. We implement custom AdaRMSNorm and fused residual-add plus RMSNorm kernels. In $\pi_{0.5}$, we produce the Q, K, and V projections with one GEMM, combine the MLP gate and up projections in another, and fuse the GeGLU or SwiGLU activation with its multiplication. MiniCPM-Robot provides a different example. Its Qwen3.5 hybrid backbone uses one Triton kernel for RMSNorm with the SiLU gate and another for depthwise causal Conv1d, SiLU, and the Q/K/V split. The latter writes Q, K, and V directly to three contiguous outputs instead of materializing and rereading a combined activated tensor. On H20 in BF16, adding these kernels to the CUDA Graph path raises throughput from 33.28 to 36.77~Hz, a further 10.5\% gain. We check both kernels against their PyTorch references in FP32 and BF16.

\paragraph{Graph replay.}
Iterative policies often repeat the same shapes and control flow while only a small part of the state changes. We let each model path declare the validity scope of reusable state, while the model runner owns the corresponding tensors and memory. In the standard $\pi_{0.5}$ configuration, the vision-language prefix remains fixed across ten Euler steps. We compute it once and retain its KV tensors, attention metadata, index arrays, and fixed buffers while the action tokens change. We also construct the sinusoidal timestep table and each layer's AdaRMS modulation table before denoising. Once the buffers and control flow are stable, the runner captures the full loop in the matching CUDA Graph bucket. A new observation or condition invalidates the prefix state before the next prediction.

\paragraph{Profiled operator selection.}
We do not bind a Layer to one kernel implementation because the fastest choice depends on the tensor shape, data type, accelerator, and execution configuration. The operator selector applies rules over these inputs and dispatches to \textcolor{phyaiblue}{\texttt{phyai-kernel}}, a framework operator, or a library kernel. The $\pi_{0.5}$ action expert, for example, issues a 50-token query against a longer prefix with head dimension 256. For this shape, we profile FlashInfer~\citep{flashinfer} FA2 prefill as faster than the automatically selected FA3 path and record FA2 in the hardware profile. The runner allocates its workspace during setup and refreshes the attention plan once per action chunk before graph replay. Another accelerator can select a different implementation without changing the modeling code.

\paragraph{Quantization.}
We quantize models offline with round-to-nearest (RTN) through \textcolor{phyaiblue}{\texttt{phyai-model-optimizer}}. At execution time, we use high-performance operators implemented by FlashInfer and Humming~\citep{flashinfer,humming}. We currently support W4A16, W8A16, W8A8, and W4A8. The phase and batch measurements in this report use BF16, so the reported speedups do not depend on these quantized paths.

\subsection{Parallel Execution}

We specify the data-parallel (DP), tensor-parallel (TP), and classifier-free-guidance (CFG) degrees in the execution configuration for each model and deployment. The scheduler forms the corresponding device groups and assigns model runners and requests to them. Each model runner owns the request state local to its rank, while stateless modeling modules use the same interface for single- and multi-GPU execution. Distributed Layers implement the sharded operators and collectives required by the configured groups.

With DP, the scheduler assigns independent requests to separate model runners. Each runner owns its buffers, caches, and solver state and executes the same single-GPU modeling and Layers path. DP workers do not exchange state while processing these requests. The $\pi_{0.5}$ implementation uses this arrangement, so adding replicas does not create a separate distributed model path.

Cosmos3-Nano-Policy-DROID uses TP and CFG parallelism for different parts of one request. Its GEN transformer jointly updates the video and action tokens, and TP partitions this computation within each branch. The conditional and unconditional CFG branches remain independent until guidance, so the scheduler runs them in separate TP groups. For this two-branch CFG path, the ranks form a $2\times N_{\mathrm{TP}}$ mesh. A rank coordinate $(c,t)$ identifies the CFG branch $c$ and TP shard $t$, with $c=0$ for the conditional branch and $c=1$ for the unconditional branch.

Within each TP group, the QKV and MLP gate and up projections are column parallel, and the attention heads are partitioned across ranks. The attention-output and MLP-down projections are row parallel, and Distributed Layers combine their partial outputs with all-reduce. Video and action tokens remain in the same GEN sequence, so one transformer forward produces both velocities.

\begin{figure}[H]
    \centering
    \includegraphics[width=\linewidth]{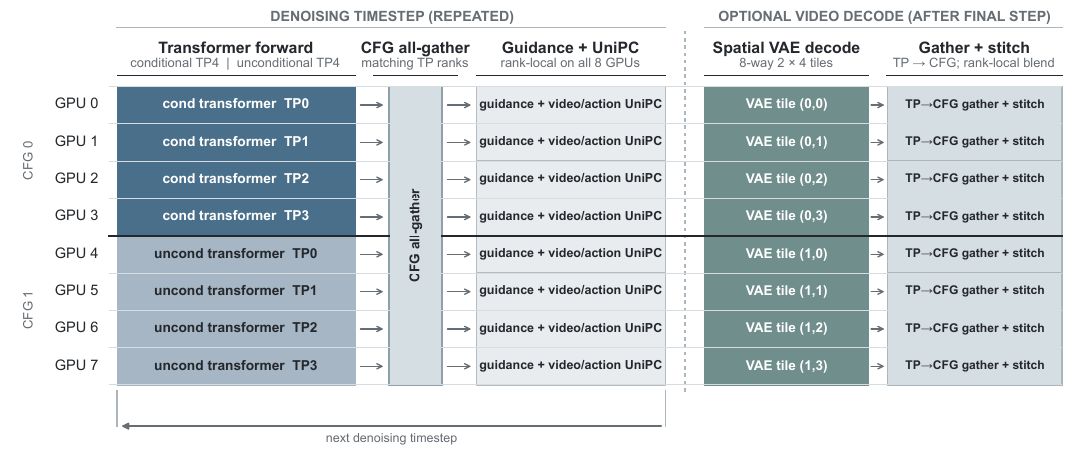}
    \caption{Eight-GPU Cosmos3 execution with two CFG branches, TP4 within each branch, and optional tiled video decoding.}
    \label{fig:cosmos-parallelism}
\end{figure}

After both branches complete a denoising step, matching TP ranks execute a CFG all-gather to obtain the conditional and unconditional velocities. Each rank then applies

\begin{equation}
    v_{\mathrm{guided}}=v_{\mathrm{uncond}}+\gamma
    \left(v_{\mathrm{cond}}-v_{\mathrm{uncond}}\right)
\end{equation}

to the video and action velocities, where $\gamma$ is the guidance scale and is independent of the CFG parallel degree. The model runner on each rank advances its local video and action UniPC solver states before the next denoising step. In the eight-GPU configuration, the scheduler uses two CFG groups with TP degree four. Ranks 0 to 3 form the conditional TP4 group, and ranks 4 to 7 form the unconditional group. The CFG all-gather pairs are $(0,4)$, $(1,5)$, $(2,6)$, and $(3,7)$.

Action-only serving ends after the final solver update. If the request also includes video decoding, the scheduler reuses the eight ranks for a spatially tiled VAE. The VAE weights are replicated, and each rank decodes one tile of a row-major $2\times4$ grid before the runtime gathers and stitches the tiles.

\FloatBarrier

%% file: 04-results.tex
\section{Real-Time at the Edge and Scale in the Cloud}
\label{sec:experiments}
\suppressfloats[t]

\subsection{Measurement Settings}
\label{sec:measurement-settings}

We report two sets of measurements. The first compares single-request latency across the models and devices available to us. The second combines static-batch sweeps with phase profiles for PI0.5, GR00T, and Cosmos3. PI0.5 and Cosmos3 use a Hopper-series GPU and Thor, while the GR00T profile uses an RTX~5090. Together, these results cover local inference and the batch behavior relevant to shared edge and cloud execution.

Unless stated otherwise, latency is the model-runner time required to produce one complete action chunk. In the notation of \Cref{eq:critical-latency}, this measurement is our proxy for $L_{\mathrm{inference}}$, not the full $L_{\mathrm{critical}}$. It excludes observation capture, transport to and from the runner, request queueing, and actuator handoff. The PI0.5 and Cosmos3 profiles use CUDA events after warm-up and exclude model loading. We compute speedup by dividing official-path latency by PhyAI latency for the same model and device. The benchmark harnesses are available with the project.\footnote{\href{https://github.com/mingti-org/phyai/tree/main/benchmark}{github.com/mingti-org/phyai/tree/main/benchmark}}

\begin{table}[tbp]
    \centering
    \footnotesize
    \begin{tabularx}{\linewidth}{@{}L{0.21\linewidth}L{0.23\linewidth}Y@{}}
        \toprule
        Study & Hardware & Configuration \\
        \midrule
        Single-request latency & RTX~5090; Thor; A40; H100; H20 $\times 8$ & $\pi_0$, $\pi_{0.5}$, GR00T N1.7, MiniCPM-Robot, and Cosmos3; available runtimes; FP8 points marked explicitly \\
        PI0.5 batch and phase analysis & Hopper-series GPU; Thor & BF16; batch sizes 1 to 32; three cameras; 50 actions; ten Euler steps; CUDA Graphs \\
        GR00T batch and phase analysis & RTX~5090 & BF16; batch sizes 1 to 32; two $256\times256$ cameras; action shape $40\times132$; four flow-matching steps; CUDA Graphs \\
        Cosmos3 batch and phase analysis & Hopper-series GPU; Thor & BF16; batch sizes 1 to 16; 33 frames at $480\times832$; 32 actions; four denoising steps; CFG 3.0; eager execution \\
        \bottomrule
    \end{tabularx}
    \caption{Measurement settings and configurations used in this section.}
    \label{tab:evaluation-matrix}
\end{table}

\FloatBarrier

For the batch analysis, we time the execution phases exposed by each model runner. In PI0.5, we measure the vision tower, language prefix, and repeated action expert separately. For GR00T, we measure Prepare, Backbone, and Action Head; Backbone contains the vision tower and selected text-backbone layers, while Action Head contains VL refinement, state encoding, request-local cross-attention K/V projections, and the four-step action denoising loop. In Cosmos3, we separate the condition phase from repeated generation. These phase names follow each model's execution path and are not intended as a one-to-one decomposition across architectures.

For PI0.5 and Cosmos3, $T_i(B)$ is measured phase time, $F_i$ is architecture-derived FLOPs per sample, and $Q_i(B)$ is modeled logical tensor traffic. We report achieved throughput and logical arithmetic intensity as

\begin{equation}
    P_i(B)=\frac{B F_i}{T_i(B)},
    \qquad
    I_i^{(\mathrm{logical})}(B)=\frac{B F_i}{Q_i(B)},
\end{equation}

with

\begin{equation}
    Q_i(B)=Q_{i,w}(B)+Q_{i,a}(B)+Q_{i,\mathrm{KV}}(B)+Q_{i,\mathrm{I/O}}(B).
\end{equation}

BF16 tensors use two bytes per element. $Q_i(B)$ covers weights, activations, K/V cache, and boundary I/O along the benchmarked path. It excludes setup-only operations and unmaterialized attention scores. We use this intensity to construct a device-specific reference envelope,

\begin{equation}
    P_i^{(\mathrm{ref})}(B)=\min\!\left(P_{\mathrm{peak}}, I_i^{(\mathrm{logical})}(B)B_{\mathrm{memory}}\right),
    \label{eq:hardware-roofline}
\end{equation}

where $P_{\mathrm{peak}}$ comes from a local $8192\times8192$ BF16 GEMM and $B_{\mathrm{memory}}$ comes from a 4~GiB device-copy benchmark. We take the best of 50 iterations after ten warm-up iterations for both roofs. Phase time comes from GPU-side profiler scopes, and the FLOP counts come from model structure and execution frequency rather than hardware counters. The profiles use PyTorch~2.11.0 and CUDA~13.0.

Logical tensor traffic is not measured DRAM traffic. Cache residency and fusion may reduce HBM transfers, while reloads, tiling, and workspaces may increase them. The envelope is a phase-level diagnostic, and points below it may reflect small shapes or launch overhead. We use it to compare phases, batches, and devices within these runners, not to assess distributed scaling.

\FloatBarrier
\subsection{Single-Request Latency}
\label{sec:edge-results}

\Cref{tab:single-request-summary} lists every official-to-PhyAI pair in the comparison. On RTX~5090, PhyAI is $4.05\times$ faster for $\pi_0$, $1.82\times$ for $\pi_{0.5}$, $2.28\times$ for GR00T, and $2.02\times$ for MiniCPM-Robot. On Thor, the speedups are $2.64\times$, $1.67\times$, $1.40\times$, and $2.78\times$, respectively. We also measured a $4.31\times$ speedup for GR00T on A40, $4.65\times$ for MiniCPM-Robot on H100, and $2.08\times$ for Cosmos3-Nano-Policy-DROID on eight H20 GPUs with CFG=2 and TP=4.

\begin{table}[tbp]
    \centering
    \footnotesize
    \begin{tabular*}{\linewidth}{@{\extracolsep{\fill}}L{0.24\linewidth}L{0.18\linewidth}>{\raggedleft\arraybackslash}p{0.14\linewidth}>{\raggedleft\arraybackslash}p{0.14\linewidth}>{\raggedleft\arraybackslash}p{0.14\linewidth}@{}}
        \toprule
        Model & Device & Official & PhyAI & Speedup \\
        \midrule
        $\pi_0$ & Thor & 321.030~ms & 121.500~ms & $2.64\times$ \\
        $\pi_0$ & RTX~5090 & 127.480~ms & 31.475~ms & $4.05\times$ \\
        $\pi_{0.5}$ & Thor & 179.503~ms & 107.699~ms & $1.67\times$ \\
        $\pi_{0.5}$ & RTX~5090 & 52.076~ms & 28.613~ms & $1.82\times$ \\
        GR00T N1.7 & Thor & 133.467~ms & 95.366~ms & $1.40\times$ \\
        GR00T N1.7 & RTX~5090 & 46.962~ms & 20.623~ms & $2.28\times$ \\
        GR00T N1.7 & A40 & 199.703~ms & 46.348~ms & $4.31\times$ \\
        MiniCPM-Robot & Thor & 156.207~ms & 56.274~ms & $2.78\times$ \\
        MiniCPM-Robot & RTX~5090 & 47.199~ms & 23.330~ms & $2.02\times$ \\
        MiniCPM-Robot & H100 & 105.380~ms & 22.640~ms & $4.65\times$ \\
        Cosmos3 Nano Policy & H20 $\times 8$ & 2460.000~ms & 1180.000~ms & $2.08\times$ \\
        \bottomrule
    \end{tabular*}
    \caption{We measure single-request latency relative to the official paths; lower values indicate faster inference.}
    \label{tab:single-request-summary}
\end{table}

The specialized-runtime results in \Cref{fig:speedup-summary} are more mixed. The two PI0 FlashRT measurements use FP8, with 49.5~ms on Thor and 19.7~ms on RTX~5090, so they are not precision-matched to the corresponding PhyAI results. On RTX~5090, realtime-vla, FlashRT, and PhyAI run PI0.5 in 27.5, 28.5, and 28.6~ms, respectively. FlashRT is also faster than PhyAI for GR00T on Thor, RTX~5090, and A40. The measured advantage is therefore relative to the official paths, not every specialized runtime.

For MiniCPM-Robot, the official baseline is the OpenBMB/MiniCPM-Robot implementation.\footnote{\href{https://github.com/OpenBMB/MiniCPM-Robot}{github.com/OpenBMB/MiniCPM-Robot}} The PhyAI result uses the PhyAI FLA (Flash Linear Attention) backend \citep{yang2024fla}. For Cosmos3-Nano-Policy-DROID, the official path takes 2.46~s and PhyAI takes 1.18~s on H20 $\times 8$ with CFG=2 and TP=4. FlashRT now supports Cosmos3, but we do not have a matched FlashRT latency for this configuration.

\FloatBarrier
\subsection{Batch Scaling and Phase Analysis}
\label{sec:cloud-results}
\label{sec:phase-analysis}

Batching can raise accelerator utilization, but each request must wait for the full batch to complete. We report both throughput and full-batch latency, then use the phase profiles defined in \Cref{sec:measurement-settings} to explain the observed scaling. The measurements use static batches and do not include queueing or network delay.

\paragraph{PI0.5.}

\begin{figure}[tbp]
    \centering
    \includegraphics[width=0.96\linewidth]{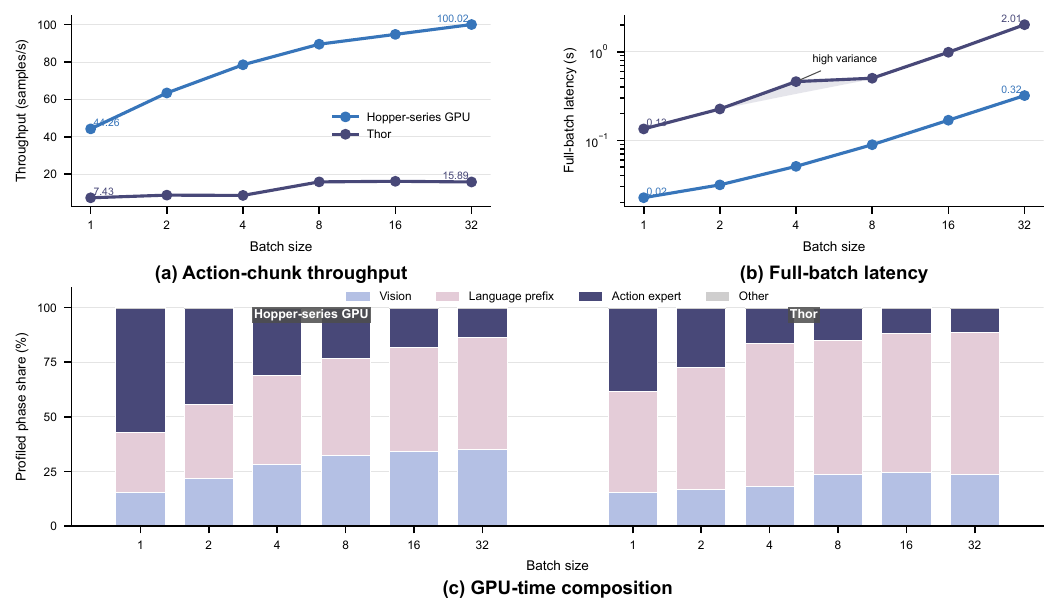}
    \caption{PI0.5 BF16 batch scaling with CUDA Graphs. \textbf{(a)}, Action-chunk throughput. \textbf{(b)}, Mean full-batch latency; shading marks the Thor batch-four range. \textbf{(c)}, GPU-time share by phase.}
    \label{fig:pi05-batch-scaling}
\end{figure}

\Cref{fig:pi05-batch-scaling} shows the BF16 PI0.5 sweep with CUDA Graphs, three camera views, a 50-action chunk, and ten Euler steps. Each point follows ten warm-up iterations and 50 timed iterations. On a Hopper-series GPU, batch size one takes 22.59~ms and processes 44.26 samples/s. Batch size 32 reaches 100.02 samples/s, but the synchronized batch completes in 319.94~ms. Its amortized cost is 10.00~ms per sample. A rollout worker can use the 100.02 samples/s capacity, while a robot waits for the 319.94~ms batch to finish.

Thor reaches its plateau earlier. Throughput rises from 7.43 samples/s at batch size one to 16.25 samples/s at batch size 16, then falls to 15.89 samples/s at batch size 32. The batch-size-four point is noisy: we measured a mean of 459.8~ms, a standard deviation of 56.2~ms, and a range of 334.2 to 489.7~ms.

\paragraph{GR00T.}

\begin{figure}[!tbp]
    \centering
    \includegraphics[width=0.96\linewidth]{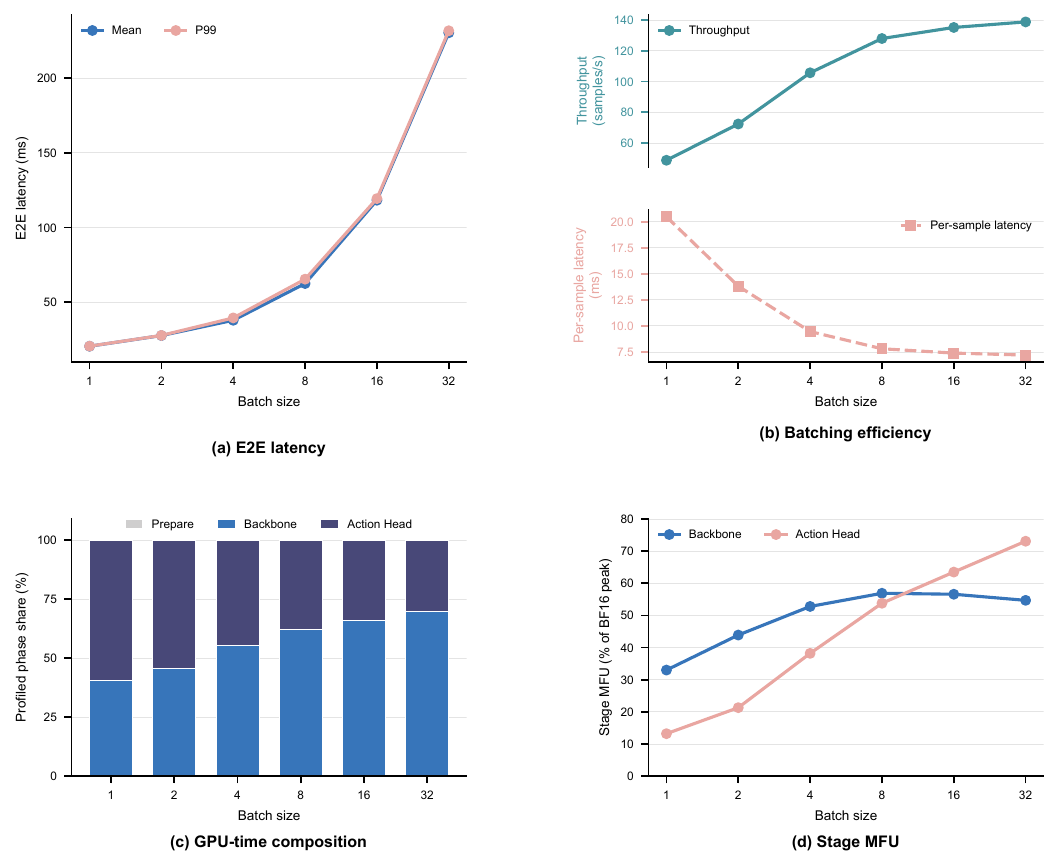}
    \caption{GR00T-N1.7 BF16 batch profile on RTX~5090. \textbf{(a)}, Mean and P99 end-to-end latency. \textbf{(b)}, Throughput and per-sample latency. \textbf{(c)}, GPU-time share for Prepare, Backbone, and Action Head. \textbf{(d)}, Phase MFU relative to the measured BF16 peak. Each batch uses eight warm-up iterations, followed by 30 timed iterations for end-to-end latency and 10 CUDA Event windows for phase measurements.}
    \label{fig:groot-batch-profile}
\end{figure}

\Cref{fig:groot-batch-profile} shows the BF16 GR00T-N1.7 profile on an RTX~5090. The end-to-end mean latency grows from 20.49~ms at batch size one to 230.50~ms at batch size 32, while throughput increases from 48.8 to 138.8 samples/s. Batch size eight already reaches 92.2\% of the batch-size-32 throughput, so larger batches mainly increase the synchronized wait. The corresponding per-sample cost falls from 20.491 to 7.203~ms.

The phase balance changes with batch size. At batch size one, Action Head accounts for 59.2\% of profiled time and Backbone for 40.7\%. Backbone becomes the larger phase at batch size four and reaches 69.7\% at batch size 32. Backbone MFU rises from 33.0\% to 56.9\% by batch size eight and ends at 54.7\%, whereas Action Head MFU rises from 13.2\% to 73.1\%. Without a reconstructed activation and cache account, we do not assign either phase a logical arithmetic intensity or ridge classification. At batch size 32, Backbone reaches 121.3~TFLOPS and Action Head reaches 162.1~TFLOPS.

\paragraph{Cosmos3.}

\begin{figure}[!tbp]
    \centering
    \includegraphics[width=0.96\linewidth]{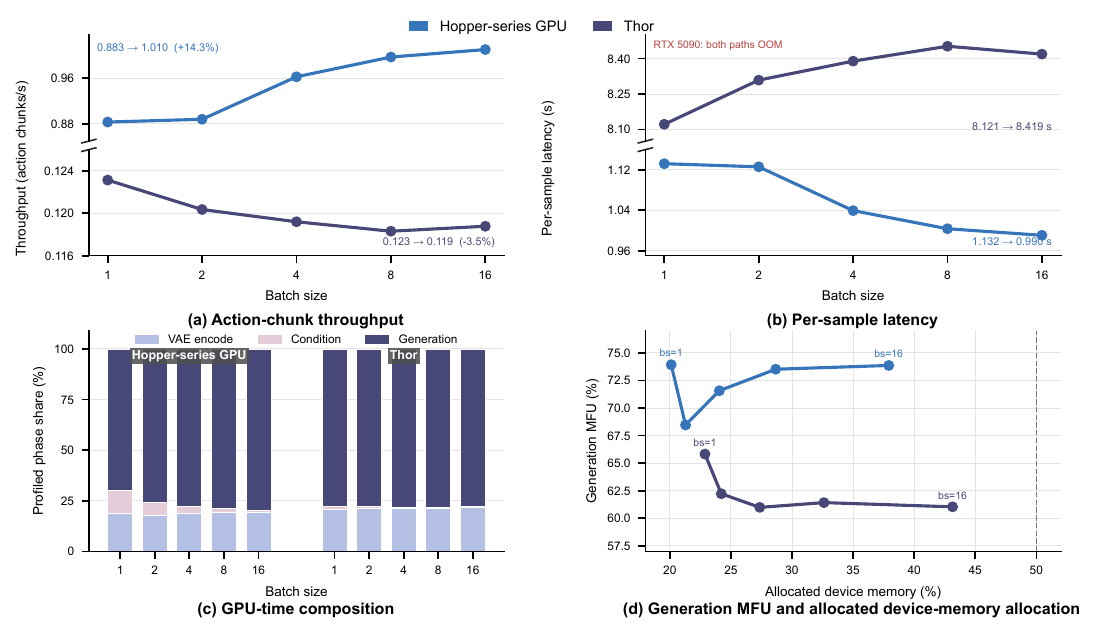}
    \caption{Cosmos3 BF16 batch profile. \textbf{(a)}, Action-chunk throughput. \textbf{(b)}, Per-sample latency. \textbf{(c)}, GPU-time share by phase. \textbf{(d)}, Generation MFU and allocated memory. Panels \textbf{(a)} and \textbf{(b)} use broken axes; RTX~5090 ran out of memory.}
    \label{fig:cosmos-latency}
\end{figure}

\Cref{fig:cosmos-latency} shows the corresponding BF16 sweep for Cosmos3. The configuration uses four denoising steps, CFG scale 3.0, a 32-action chunk at 15~Hz, and 33 frames at $480\times832$. CUDA Graphs are disabled in this eager runner. Each point is the mean of five timed iterations after two warm-ups and excludes optional rollout-video decode. On a Hopper-series GPU, batch size one takes 1.132~s and yields 0.883 chunks/s. Batch size 16 takes 15.85~s for the full batch and yields 1.010 chunks/s, only 14.3\% more throughput. On Thor, throughput changes from 0.123 to 0.119 chunks/s.

\FloatBarrier

\begin{figure}[!tbp]
    \centering
    \includegraphics[width=0.96\linewidth]{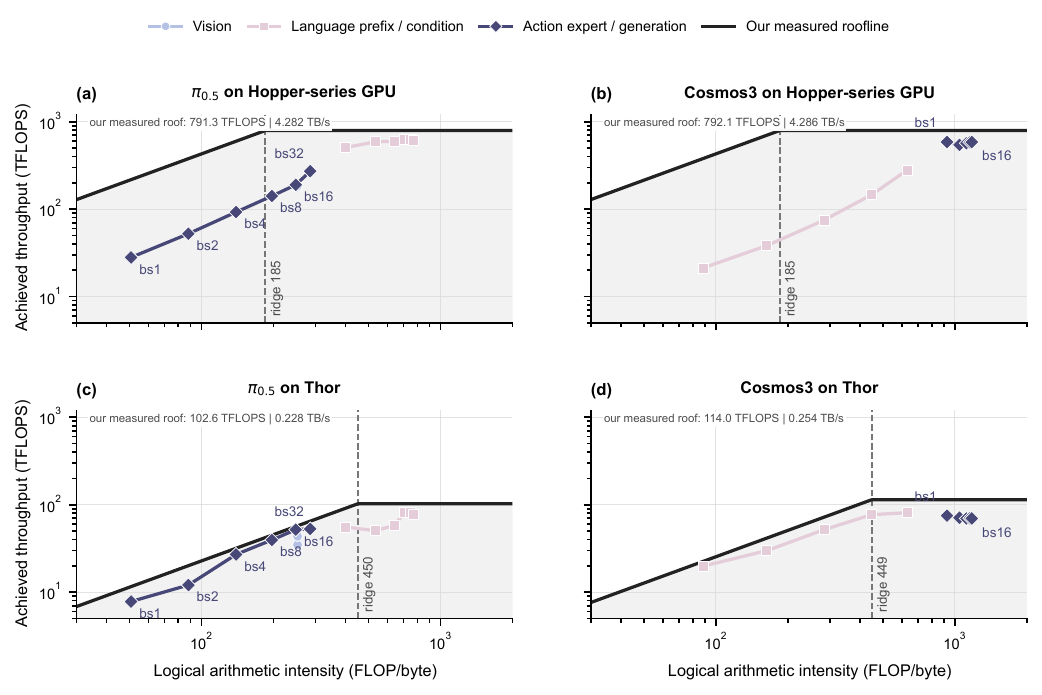}
    \caption{BF16 Rooflines for PI0.5 and Cosmos3 using logical arithmetic intensity on a Hopper-series GPU and Thor.}
    \label{fig:phase-roofline}
\end{figure}

The Cosmos3 profile in \Cref{fig:phase-roofline} falls in a different regime. At batch size one, generation has a logical arithmetic intensity of 926.2~FLOP/byte and lies to the right of both ridges. It reaches 585.5~TFLOPS on the Hopper-series GPU and 75.0~TFLOPS on Thor, with MFU between 68.5\% and 73.9\% and between 61.0\% and 65.8\%, respectively. Condition crosses the Hopper ridge between batches two and four and approaches the Thor ridge at batch eight. Allocated memory remains below 44\%.

These measurements identify generation compute, rather than device-memory capacity, as the limit in the measured Cosmos3 runner. The generation phase already supplies enough work at batch size one to use the compute path heavily, so a larger batch adds little throughput and increases request latency. This result motivates within-request parallelism when lower Cosmos3 latency is required, but the single-GPU profile does not measure TP or CFG scaling.

\FloatBarrier

\Cref{fig:phase-roofline} explains why the two batch sweeps behave differently. At batch size one on the Hopper-series GPU, the PI0.5 action expert accounts for only 8.8\% of estimated FLOPs but 57.2\% of profiled time. Its repeated small GEMMs reach 28.1~TFLOPS at a logical arithmetic intensity of 50.8~FLOP/byte, to the left of the measured ridge at 184.8~FLOP/byte. The combination of low achieved throughput and a large time share identifies the action loop as the low-batch bottleneck.

PI0.5's bottleneck changes with batch size. On the Hopper-series GPU, the expert crosses the ridge between batches four and eight and reaches 271.3~TFLOPS at batch size 32 against the 791.3~TFLOPS compute roof. Its time share falls to 13.5\%, while language and vision account for 51.4\% and 35.1\%. On Thor, language crosses between batches one and two, but vision and expert remain left of the ridge through batch 32. Language's share rises from 46.1\% to 64.9\%, so expert batching helps only until the vision-language path dominates.

\FloatBarrier
\input{04-subsection-cloud-rl-rollout}

\subsection{Summary}

PhyAI is faster than the official path in all 11 measured pairs, with speedups from $1.40\times$ to $4.65\times$. This comparison is not fully precision-matched, and specialized runtimes are faster in several configurations. The batch sweeps show three different regimes: PI0.5 gains throughput until its vision-language path becomes dominant, GR00T reaches $92.2\%$ of its batch-size-32 throughput by batch size eight while its bottleneck shifts from Action Head to Backbone, and Cosmos3 gains little because generation remains on the compute side of the Roofline.

These measurements cover GPU execution time and static batches. They do not include observation capture, transport to and from the runner, request queues, actuator handoff, tail latency, integrated RL rollout, or closed-loop task quality. We therefore use them to characterize runtime performance, not to claim an end-to-end improvement in robot success or safety.

\FloatBarrier

%% file: 04-subsection-cloud-rl-rollout.tex
\subsection{Cloud RL Rollout}

RL post-training is used to align LLMs and VLMs and improve their performance on downstream tasks. Frameworks such as verl and slime coordinate trajectory generation with distributed policy updates.\footnote{Official projects: \href{https://github.com/verl-project/verl}{verl} and \href{https://github.com/THUDM/slime}{slime}.} Because each update consumes trajectories generated during rollout, policy-inference latency directly affects wall-clock training time. These frameworks therefore use dedicated rollout backends such as vLLM and SGLang instead of relying on the training stack itself for generation \citep{vllm,sglang}.

In contrast, specialized rollout infrastructure for VLA RL remains limited. RLinf orchestrates distributed environment interaction and trajectory collection while delegating policy inference to a configurable backend. Its default Hugging Face backend executes $\pi_{0.5}$ through the reference PyTorch path rather than a VLA-specialized inference runtime. PhyAI can serve policy-inference requests at this backend boundary, while RLinf retains environment orchestration and policy optimization.

We profiled $\pi_{0.5}$ GRPO training on LIBERO-10 on a single node with four NVIDIA A800 80~GB GPUs and no NVLink. Starting from RLinf's default LIBERO-10 recipe, we reduced the number of parallel environments from 64 to 32, the actor micro-batch size from 128 to 64, and the global batch size from 2,048 to 1,024 to fit the available hardware. We disabled evaluation, checkpoint writing, and video capture to isolate rollout and actor training. \Cref{tab:cloud-rl-rollout-config} lists the profiled configuration.

\begin{table}[H]
    \centering
    \footnotesize
    \begin{tabularx}{\linewidth}{@{}L{0.56\linewidth}Y@{}}
        \toprule
        Setting & Value \\
        \midrule
        Training environments & 32 \\
        Steps per rollout epoch & 480 \\
        Environment group size & 4 \\
        Rollout epochs & 8 \\
        Executed actions per policy query & 10 \\
        Flow-matching denoising steps & 4 \\
        Training micro batch size & 64 \\
        Training global batch size & 1,024 \\
        Actor update epochs & 4 \\
        Cluster nodes & 1 \\
        Component placement & all \\
        Rollout pipeline stages & 1 \\
        \bottomrule
    \end{tabularx}
    \caption{Configuration of the profiled $\pi_{0.5}$ GRPO run on LIBERO-10.}
    \label{tab:cloud-rl-rollout-config}
\end{table}

\begin{figure}[H]
    \centering
    \includegraphics[width=\linewidth]{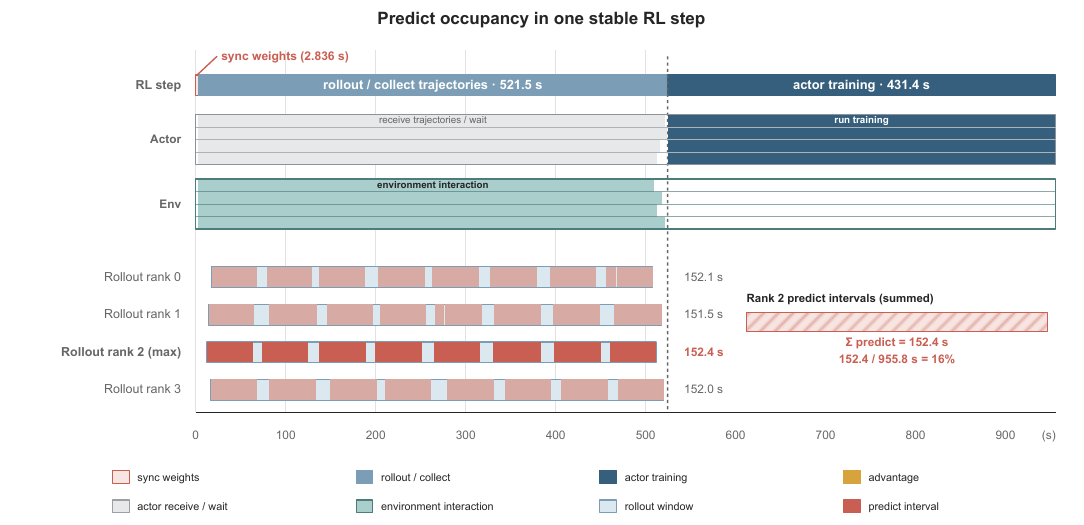}
    \caption{Timeline of one stable RL step on four A800 80~GB GPUs.}
    \label{fig:rl-step-predict-occupancy}
\end{figure}

\Cref{fig:rl-step-predict-occupancy} shows a representative steady-state RL step measured after initialization and one warm-up step. We extracted wall-clock durations from NVTX ranges in Nsight Systems traces and aligned the actor, environment, and rollout ranks to the same RL-step window. For each rank, the \texttt{predict} duration is the sum of its \texttt{predict} intervals within that window.

The step took 955.8~s: weight synchronization took 2.836~s, rollout and trajectory collection took 521.5~s, and actor training took 431.4~s. The accumulated per-rank \texttt{predict} time ranged from 151.5 to 152.4~s. Because the rollout ranks execute concurrently, the maximum per-rank duration, rather than their sum, determines the critical-path contribution. Rank~2 took 152.4~s, accounting for 15.9\% of the RL-step wall-clock time and 29.2\% of the rollout stage.

Under the same configuration, PhyAI achieved a measured $2.55\times$ speedup over RLinf in mean \texttt{predict} time. We estimate end-to-end performance using an idealized Amdahl-law model that holds all non-\texttt{predict} work constant, keeps the accelerated calls on the critical path, and assumes no integration overhead. With \texttt{predict} accounting for 15.9\% of the critical path, the projected step time decreases from 955.8~s to 863.2~s, corresponding to a 9.7\% reduction and a $1.11\times$ end-to-end speedup.

This projection does not represent a measured end-to-end speedup. Instead, it characterizes the opportunity as rollout becomes more inference-intensive. We further observed that two changes increased the fraction of step time spent in \texttt{predict}: increasing the policy-inference workload through additional flow-matching denoising steps, and reducing actor update epochs to shorten training. Under these configurations, the \texttt{predict} share could reach 40\%--50\%. If PhyAI maintains its measured $2.55\times$ speedup, the idealized Amdahl-law estimate gives a 24.3\%--30.4\% reduction in step time and a $1.32\times$--$1.44\times$ end-to-end speedup.

PhyAI's projected end-to-end speedup therefore grows as policy inference occupies a larger fraction of the rollout critical path. For VLA policies with iterative action generation, this fraction can become substantial, so reducing policy-inference time can shorten RL training.

\FloatBarrier

%% file: 05-conclusion-discussion.tex
% note(wangchenghua): Section 7 still needs another writing pass.
\section{Future Work}
\label{sec:future-work}
\suppressfloats[t]

\subsection{Mega Kernels}

At batch size one on the Hopper-series GPU, the PI0.5 action expert accounts for 8.8\% of estimated FLOPs but 57.2\% of profiled time. Its ten-step loop still launches many small kernels. MPK lowers tensor programs to SM-level task graphs and uses decentralized scheduling inside one persistent kernel to pipeline work across operators \citep{cheng2026mpk}. Event Tensor represents dependencies between tiled tasks and supports shape- and data-dependent execution through static and dynamic scheduling \citep{jin2026event}. PhyAI could use these abstractions to compile the repeated expert path as one scheduled graph while preserving the current fallback. Evaluation should measure end-to-end action-chunk latency and numerical agreement.

\subsection{Climbing the Performance Mountain with Kernel Agents}

Thor still has few architecture-specific kernels, so many common shapes fall back to framework or library implementations. Recent kernel agents make it practical to tune these paths systematically. Following AutoKernel, we will profile each model on Thor, rank hotspots by GPU-time share, and extract the leading operators or fused regions as standalone CUDA or Triton targets. Each candidate goes through the same correctness and timing harness. The harness discards regressions, and the next target is chosen by its estimated effect on end-to-end latency \citep{autokernel}.

For each target, we will use AVO's agentic variation loop rather than a fixed set of mutation rules \citep{chen2026avo}. The agent reads the candidate's lineage, Thor-specific notes, and compiler and benchmark feedback before proposing or repairing an edit. The search can change tiling, fusion boundaries, data layout, and memory staging. A candidate enters \textcolor{phyaiblue}{\texttt{phyai-kernel}} only after numerical checks and shape sweeps; unsupported shapes keep the existing fallback. We will report search cost and end-to-end model latency alongside kernel speedup.

\subsection{PhyAI as an RLinf Rollout Backend}

RLinf separates rollout orchestration from policy execution through an inference-backend interface \citep{rlinf}. PhyAI already follows this interface, but our current rollout result comes from simulation. We will make PhyAI a usable RLinf backend. RLinf will keep environment and training orchestration, while PhyAI batches policy calls from rollout workers and executes them through its optimized single- or multi-GPU paths. This should shorten the inference portion of each rollout step. We will measure full rollout-step latency and training throughput, with separate timings for policy calls, queueing, and backend synchronization.

\subsection{Broader Model and Accelerator Support}

Our next model targets include Xiaomi-Robotics-1, Xiaomi-Robotics-U0, LingBot-VA 2.0, LingBot-VLA 2.0, HY-Embodied-0.5, and Wall-OSS-0.5 \citep{xiaomi-robotics-1,li2026xiaomiroboticsu0,lingbotva-2,lingbot-vla,team2026hy,yu2026wall}. We will place each model's preprocessing, conditioning, and action path behind the existing adapter interface before adding model-specific graph and kernel optimizations. Hardware targets include Ascend 910/950, Kunlunxin P800, Zhenwu M890, AMD Embedded X100, and AMD Instinct MI355X for edge and datacenter deployment. For each supported model-device pair, we will report functional coverage and end-to-end latency.

\subsection{A Production-Grade Inference Serving Protocol}

Robot model-as-a-service (MaaS) lacks a common serving protocol comparable to the OpenAI API. A request schema would only add another format. Robot inference is a timed, stateful stream: observations arrive continuously, actions expire, and newer state can replace in-flight work. Like a video codec, the protocol should exploit temporal redundancy with keyed observations and delta updates. It must also define action horizons, cache reuse, deadlines, backpressure, and recovery. We will prototype it between robot clients and PhyAI servers.

% note(wangchenghua): Section 8 still needs another writing pass.
\section{Discussion and Conclusion}
\label{sec:discussion}
\suppressfloats[t]

\subsection{Using the Control-Time Roofline}

The control-time Roofline gives runtime optimization a practical stopping point. When inference is slower than environment execution, reducing inference time raises the ideal control rate. Once inference fits inside the action window, the controller gets no faster under ideal overlap. \textcolor{phyaiblue}{\textbf{Once inference fits inside the action window, the remaining margin can support a larger model or a cheaper, slower accelerator, as long as the complete request still reaches the next action handoff on time.}}

For the four LIBERO suites, measured $\pi_{0.5}$ inference is already shorter than simulator environment execution. These points leave room to trade runtime speed for model capacity or hardware cost, but our data do not quantify either trade. The next experiment should sweep model size and device class under a fixed control budget, then report task success, p99 control-time margin, deadline misses, and deployment cost. This comparison would show how much of the action window can be used without giving up the required reliability.

\subsection{MaaS for Factories}

We do not treat cloud and edge deployment as competing bets. Equipping 100 robots with 100 onboard accelerators may cost more than serving the same fleet from a smaller pool of datacenter GPUs, especially when policy requests are intermittent and the onboard devices sit idle. A remote cloud service avoids that replication, but its wide-area network path adds latency and variation to the control budget. \textcolor{phyaiblue}{\textbf{For factory fleets, a local MaaS deployment can combine shared accelerators with lower network latency than a remote cloud path.}}

A factory-local service still adds network transfer and queueing, and a shared server creates a larger failure domain. Its value should be measured rather than assumed. A deployment study should compare total hardware cost, accelerator utilization, p99 critical-path latency, and deadline misses for onboard, local MaaS, and remote cloud execution under the same workload. PhyAI keeps the model path unchanged across these placements so that this choice can follow the control budget instead of requiring another inference implementation.

\subsection{Conclusion}
\label{sec:conclusion}

We therefore want one runtime that can follow a model wherever it needs to run. The same model path should work for robot MaaS and cloud RL rollouts, as well as edge-cloud and fully onboard execution. Model adapters keep its behavior intact, while the runtime adjusts batching, parallelism, kernels, and placement for each setting. Training, serving, and deployment can then share one implementation without being forced into one execution policy.

Better infrastructure gives algorithm researchers room to ask bolder questions. Time saved in execution can return as quicker control or be spent on more capable models and richer observations. We hope to build PhyAI with the community as shared infrastructure for Physical AI, and together help bring AGI into the physical world.

%% file: 06-appendix.tex
\appendix
\section{GR00T N1.7 Success Rate on LIBERO 10}
\label{sec:groot-libero10-success}

\paragraph{Inference server.}
The evaluation used an NVIDIA A40 GPU on Linux 5.15 with driver 610.43.02. PhyAI 0.1.0 ran with Python 3.12.11, PyTorch 2.11.0+cu130, and CUDA 13.0. The checkpoint was \texttt{GR00T-N1.7-LIBERO/}\allowbreak\texttt{libero\_10}. The simulator sent camera images, robot state, and task instructions to the inference service over ZeroMQ. The service prepared the model inputs and returned decoded actions to the simulator.

\paragraph{Simulation environment.}
The simulator used a LIBERO container pinned to a fixed image digest. The container included Python 3.8.20, LIBERO 0.1.0, MuJoCo 3.2.3, robosuite 1.4.0, NumPy 1.24.4, \texttt{pyzmq} 27.1.0, and \texttt{msgpack-numpy} 0.4.8. MuJoCo used EGL for offscreen rendering. Each observation contained two $256\times256$ images, one from the agent view and one from the wrist view, together with the end-effector position and orientation, gripper state, and language instruction. The client converted quaternion orientations to axis-angle form. The environment received seven control dimensions for position, rotation, and gripper state.

\paragraph{Task suite.}
The evaluation covered all ten LIBERO-10 tasks:
\begin{enumerate}[leftmargin=*,itemsep=1pt,topsep=3pt]
    \item Put both the alphabet soup and the tomato sauce in the basket.
    \item Put both the cream cheese box and the butter in the basket.
    \item Turn on the stove and put the moka pot on it.
    \item Put the black bowl in the bottom drawer of the cabinet and close it.
    \item Put the white mug on the left plate and put the yellow and white mug on the right plate.
    \item Pick up the book and place it in the back compartment of the caddy.
    \item Put the white mug on the plate and put the chocolate pudding to the right of the plate.
    \item Put both the alphabet soup and the cream cheese box in the basket.
    \item Put both moka pots on the stove.
    \item Put the yellow and white mug in the microwave and close it.
\end{enumerate}

\paragraph{Evaluation protocol.}
Each task used 50 fixed initial states, for 500 episodes in total. Within each task, the episode seeds ran consecutively from 42 through 91. The batch size was one, and the simulator executed the first eight decoded actions from each model request. Each episode ran for at most 720 simulation steps, with success evaluated after every step. The official implementation and PhyAI used the same checkpoint, initial states, task order, and simulation parameters.

\paragraph{Results.}
The official NVIDIA Isaac-GR00T implementation completed 456 of 500 episodes successfully, for a success rate of 91.2\%. PhyAI completed 459 episodes successfully, for a success rate of 91.8\%.

\begin{table}[htbp]
    \centering
    \small
    \begin{tabular}{@{}lrrr@{}}
        \toprule
        Runtime & Episodes & Successes & Success rate \\
        \midrule
        NVIDIA Isaac-GR00T & 500 & 456 & 91.2\% \\
        PhyAI & 500 & 459 & 91.8\% \\
        \bottomrule
    \end{tabular}
    \caption{GR00T-N1.7 success on LIBERO-10 under matched checkpoints, initial states, task order, and simulator settings.}
    \label{tab:groot-libero10-success}
\end{table}

\FloatBarrier

\section{Reproducing PhyAI \texorpdfstring{$\pi_{0.5}$}{pi0.5} in RoboTwin}
\label{sec:pi05-robotwin-reproduction}

\paragraph{Simulation environment.}
We used the RoboTwin integration that includes the PhyAI $\pi_{0.5}$ policy. The simulator ran in a Python 3.10 environment with the ALOHA-AgileX assets and randomized demonstration configuration. SAPIEN, MPLib, Curobo, Vulkan, and ffmpeg were checked before evaluation. The simulation environment remained separate from the PhyAI runtime because the two stacks require different Python and CUDA dependencies.

The RTX~5090 setup required driver version 580 or later. Its inference runtime used Python 3.12, PyTorch 2.11, CUDA 13, and FlashInfer. Jetson Thor used an aarch64 container with Compute Capability 11.0 and the NVIDIA graphics runtime. The Thor image was checked for PyTorch, SAPIEN, MPLib, OIDN, and Vulkan support. Source code and checkpoints were mounted read-only, while runtime caches used a separate writable directory.

\paragraph{Checkpoint and normalization.}
We used the \texttt{motus-robotics/}\allowbreak\texttt{pi0.5\_robotwin2} checkpoint. The model weights were paired with the normalization statistics for the clean randomized joint-training configuration. These statistics contain the $q_{01}$ and $q_{99}$ values for both state and action. A missing or mismatched statistics file does not always stop inference, but it changes the action scale and can sharply reduce success.

\paragraph{Evaluation protocol.}
Before running full episodes, we checked that one model request produced a finite $32\times14$ action tensor. We then ran one episode end to end before starting the 100-episode evaluation. Table~\ref{tab:pi05-robotwin-protocol} lists the fixed model and control settings.

\begin{table}[htbp]
    \centering
    \small
    \begin{tabular}{@{}ll@{}}
        \toprule
        Setting & Value \\
        \midrule
        Visual input & Three RGB camera views \\
        Robot state & 14 dimensions \\
        Inference precision & BF16 \\
        Denoising steps & 10 \\
        Action horizon & 32 \\
        Executed actions per request & First 10 \\
        Control rate & 30~Hz \\
        Noise seed & 42 \\
        Task instruction & Unseen variant \\
        \bottomrule
    \end{tabular}
    \caption{Fixed settings for the PhyAI $\pi_{0.5}$ RoboTwin evaluation.}
    \label{tab:pi05-robotwin-protocol}
\end{table}

RoboTwin originally supplied the cameras in head, right, left order. The model expected head, left, right order, so the two wrist views were swapped before inference. OpenPI and PhyAI used the same evaluation manifest, which fixed the scene, instruction, and initial noise for each episode.

\paragraph{Run validation.}
A complete evaluation produced one episode record and one video per trial, together with an aggregate result. We checked the number of records and videos before reporting success. The archived run also records the source commit, GPU driver, checkpoint revision, and measured success rate.

\FloatBarrier

\section{PhyAI \texorpdfstring{$\pi_{0.5}$}{pi0.5} Evaluation Across Four LIBERO Suites}
\label{sec:pi05-libero-four-suites}

\paragraph{Evaluation setup.}
We used \texttt{vla-evaluation-harness} to run LIBERO and call a PhyAI policy server over WebSocket. The simulator used the container image distributed with the harness, while inference ran in a separate NVIDIA PyTorch container. This separation kept the LIBERO and PhyAI Python dependencies independent. On ARM64 hosts, the simulator image was built natively instead of running the x86 image through emulation. We also checked that no other job occupied the evaluation GPU before collecting timing data.

\paragraph{Model configuration.}
The server loaded a converted $\pi_{0.5}$ LIBERO checkpoint together with the PaLI-Gemma 3B tokenizer and processor. Each request contained an agent-view image, a wrist-camera image, and the robot state. Inference used BF16, FlashInfer for attention and linear layers, the PhyAI normalization kernel, and CUDA Graphs. The action chunk contained ten actions. CUDA Graph capture completed before the timed episodes began.

\paragraph{Simulation protocol.}
The evaluation covered \texttt{libero\_spatial}, \texttt{libero\_object}, \texttt{libero\_goal}, and \texttt{libero\_10}. Each suite contains ten tasks, and each task used 50 episodes, giving 500 episodes per suite and 2,000 episodes overall. The harness ran in synchronous mode and executed action chunks of size ten. All four runs used the same harness settings and simulator image.

Each episode record contained the task outcome, number of environment steps, model timing, and chunk length. We accepted a run only when both the raw and served chunks contained at most ten actions. The final check compared the number of result records with the expected episode count before aggregating success and timing.

\paragraph{Results.}
Table~\ref{tab:pi05-libero-four-suites} reports the reproduced results. PhyAI completed 1,949 of 2,000 episodes successfully, an aggregate success rate of 97.45\%. Mean model inference time remained between 36.10 and 36.33~ms across the four suites.

\begin{table}[htbp]
    \centering
    \small
    \begin{tabular}{@{}lrrrrr@{}}
        \toprule
        Suite & Episodes & Successes & Success rate & Wall time & Mean inference \\
        \midrule
        \texttt{libero\_spatial} & 500 & 489 & 97.8\% & 4198.73~s & 36.33~ms \\
        \texttt{libero\_object} & 500 & 499 & 99.8\% & 5199.23~s & 36.18~ms \\
        \texttt{libero\_goal} & 500 & 490 & 98.0\% & 3981.09~s & 36.10~ms \\
        \texttt{libero\_10} & 500 & 471 & 94.2\% & 9150.13~s & 36.21~ms \\
        \bottomrule
    \end{tabular}
    \caption{PhyAI $\pi_{0.5}$ success and timing on four LIBERO suites. Each suite contains 500 episodes and uses synchronous action chunks of size ten.}
    \label{tab:pi05-libero-four-suites}
\end{table}

\clearpage

\section{Eight GPU Data Parallel Serving with PhyAI \texorpdfstring{$\pi_{0.5}$}{pi0.5}}
\label{sec:pi05-eight-gpu-demo}

\paragraph{System configuration.}
The demonstration ran on Ubuntu 22.04 with eight NVIDIA H20 GPUs, each with about 96~GB of memory. The host had 192 CPU cores and about 1.8~TiB of system memory. The PhyAI server and LIBERO simulators ran in separate containers. The server loaded the converted $\pi_{0.5}$ LIBERO checkpoint and a local PaLI-Gemma 3B tokenizer and processor, so startup did not depend on access to the model hub.

\paragraph{Concurrent serving topology.}
We launched 32 LIBERO clients: eight shards from each of the spatial, object, goal, and LIBERO-10 suites. Every client connected to rank zero of the PhyAI server. Rank zero collected at most 32 pending observations and waited up to 20~ms for a batch to fill. It then split the batch across eight data-parallel workers, with four requests assigned to each GPU, and gathered the predicted actions before returning them to the clients.

Each GPU replayed a CUDA Graph captured for a local batch of four. When fewer than 32 observations arrived within the batching window, the server padded the local inputs to the captured shape and removed the padded outputs after the gather. The graph shape therefore remained fixed even as request batches varied with simulator progress.

\begin{table}[htbp]
    \centering
    \small
    \begin{tabular}{@{}ll@{}}
        \toprule
        Setting & Demonstration value \\
        \midrule
        Data-parallel workers & 8 GPUs \\
        Concurrent LIBERO clients & 32 \\
        Maximum server batch & 32 \\
        Fixed local graph batch & 4 per GPU \\
        Maximum batching wait & 20~ms \\
        Action chunk & 1 \\
        Inference precision & BF16 \\
        Attention and linear backend & FlashInfer \\
        Normalization backend & PhyAI kernel \\
        Graph execution & CUDA Graphs \\
        \bottomrule
    \end{tabular}
    \caption{Serving configuration for the eight-GPU PhyAI $\pi_{0.5}$ LIBERO demonstration.}
    \label{tab:pi05-eight-gpu-demo}
\end{table}

\paragraph{Action handoff.}
The recorded demonstration used an action chunk of one. Each client sent a new observation after every environment step, and the server returned one action for that observation. A larger chunk would reduce request frequency but leave the simulator open loop for more steps. We kept the chunk at one to expose the behavior of the concurrent server at every action handoff.

\paragraph{Run validation.}
The simulator clients started only after all eight workers had completed CUDA Graph capture. We checked that the server reported a maximum batch of 32 and that all 32 client shards produced episode records with success and timing fields. GPU activity was monitored during the run to confirm that the data-parallel workers were active.

\clearpage

\section{Experimental setup and LIBERO evaluation of PhyAI \texorpdfstring{$\pi_0$}{pi0}}
\label{sec:pi0-libero-reproduction}

\paragraph{Environment and model inputs.}
Experiments ran on Linux with Python 3.12 or later. Repository dependencies were installed with uv from the committed lockfile. Inference used a CUDA-capable NVIDIA GPU; headless servers rendered MuJoCo 3.8.1 through EGL. The $\pi_0$ checkpoint, PaLI-Gemma tokenizer, and LIBERO assets were loaded from local storage with Hugging Face offline mode enabled. The LIBERO environments, observation preprocessing, and MuJoCo interface followed LeRobot's LIBERO extra, while PhyAI Engine provided policy inference.

\paragraph{Policy execution.}
Inference used BF16 parameters and FlashInfer paged attention, with CUDA Graphs enabled by default. The checkpoint determines whether two or three camera views are used. Each request produced a 50-step action chunk. The controller executed one action per environment step and requested another chunk only after the current chunk was exhausted. Actions were clipped to $[-1,1]$.

\paragraph{Evaluation protocol.}
We evaluated the \texttt{libero\_spatial} and \texttt{libero\_object} suites. Each suite contains ten tasks, and each task used ten episodes, giving 200 episodes in total. Evaluation used batch size one. We did not save videos or impose an additional smoke-test step limit. An episode ended when the environment returned a terminal signal or reached LIBERO's step limit. We counted an episode as successful when the environment reported success.

\paragraph{Results.}
Overall success was 143/200 (71.5\%). The rates were 71.0\% for \texttt{libero\_spatial} and 72.0\% for \texttt{libero\_object}.

\begin{table}[htbp]
    \centering
    \small
    \begin{tabular}{@{}lrrr@{}}
        \toprule
        Suite & Tasks & Episodes & Successes / episodes \\
        \midrule
        \texttt{libero\_spatial} & 10 & 100 & 71 / 100 (71.0\%) \\
        \texttt{libero\_object} & 10 & 100 & 72 / 100 (72.0\%) \\
        Overall & 20 & 200 & 143 / 200 (71.5\%) \\
        \bottomrule
    \end{tabular}
    \caption{PhyAI $\pi_0$ success on the evaluated LIBERO suites. Each task uses ten episodes.}
    \label{tab:pi0-libero-results}
\end{table}